\documentclass[12pt]{article}

\usepackage{authblk}

\usepackage[utf8]{inputenc} 
\usepackage[T1]{fontenc}    
\usepackage{hyperref}
\hypersetup{colorlinks,citecolor=blue,urlcolor=black,linkcolor=blue,linktocpage=true}
\usepackage{url}            
\usepackage{booktabs}       
\RequirePackage{natbib}
\usepackage[margin=1in]{geometry}

\usepackage{wrapfig, amsmath, amsthm, mathtools}

\usepackage{amsfonts}       
\usepackage{amssymb}
\usepackage{nicefrac}       
\usepackage{microtype}      
\usepackage{xcolor}         

\usepackage{subfigure}
\usepackage{multirow}

\usepackage{tikz, pgfplots}
\usepackage[acronym,shortcuts]{glossaries}

\pgfplotsset{compat=newest}

\def\X{{\bf X}}

\def\0{{\bf 0}}
\def\1{{\bf 1}}

\def\AM{{\mathcal A}}

\def\MM{{\mathcal M}}
\def\NM{{\mathcal N}}

\def\MM{{\mathcal M}}

\def\SM{{\mathcal S}}

\def\RB{{\mathbb R}}

\newtheorem*{remark}{Remark}
\newtheorem*{remark*}{Remark}

\newacronym{llm}{LLM}{Large Language Model}
\newacronym{camp}{CAMP}{\textbf{CA}usal \textbf{M}ethod \textbf{P}redictor}
\newacronym{scm}{SCM}{Structural Causal Model}
\newacronym{ate}{ATE}{Average Treatment Effect}

\title{Learned Causal Method Prediction}
\date{\vspace{-5ex}}

\usepackage{times}

\author[1]{Shantanu Gupta \footnote{The work was done while the author was an intern at Microsoft Research.}}
\author[2]{Cheng Zhang}
\author[2]{Agrin Hilmkil}

\affil[1]{Carnegie Mellon University}
\affil[2]{Microsoft Research}
\affil[ ]{\small{
\texttt{\href{mailto:shantang@cmu.edu}{shantang@cmu.edu}}, 
\texttt{\{\href{mailto:cheng.zhang@microsoft.com}{cheng.zhang},\href{mailto:agrinhilmkil@microsoft.com}{agrinhilmkil}\}@microsoft.com}}}

\newcommand{\name}[1]{CAMP}

\begin{document}

\maketitle

\begin{abstract}
For a given causal question, it is important to efficiently 
decide which causal inference method to use for a given dataset.
This is challenging because causal methods typically
rely on complex and difficult-to-verify assumptions,
and cross-validation is not applicable since ground truth causal quantities are unobserved.
In this work, we propose \ac{camp},
a framework for 
predicting the best method for a given dataset.
To this end, we generate datasets from a diverse set of synthetic causal models,
score the candidate methods, 
and train a model to directly predict the highest-scoring
method for that dataset.
Next, by formulating a self-supervised pre-training objective centered on dataset assumptions relevant for causal inference, we significantly reduce the need for costly labeled data and enhance training efficiency.
Our strategy learns to map implicit dataset properties to the best
method in a data-driven manner.
In our experiments, we focus on method prediction for causal discovery.
\ac{camp} outperforms selecting any individual candidate method and
demonstrates promising generalization to unseen semi-synthetic and
real-world benchmarks.
\end{abstract}

\section{Introduction}
\label{sec:intro}
Causal models are needed across diverse application domains
as they can be used to understand the underlying mechanisms behind the data
and the consequences of unseen interventions. 
While there has been sustained progress in causal inference and discovery
\citep{yao2021survey, squires2022causal}, 
the effective application of causal methods to a given dataset 
often requires a deep understanding of the available methods 
and their compatibility with the problem at hand. 
Such barriers to entry can 
preclude the widespread adoption of causal inference methods.

An exciting prospect is instructing transformer-based \acp{llm} to perform causal tasks, 
thereby granting a natural language interface for causal inference. 
A number of emergent abilities have been observed~\citep{wei2022emergent}
including some general reasoning abilities~\citep{bubeck2023sparks}. 
Unfortunately, while \acp{llm} are able to exploit domain knowledge 
relevant for causal inference~\citep{kiciman2023causal, long2023causal},
their ability to perform advanced causal reasoning
is still limited~\citep{zhang2023understanding, zevcevic2023causal}. 
Furthermore, improving causal capabilities of \acp{llm} with fine-tuning strategies,
like instruction fine-tuning~\citep{ouyang2022training},
may prove infeasible as soliciting labels 
requires expensive controlled experiments. 
Nevertheless, existing \acp{llm} have proven to be 
remarkably extensible by using external tools~\citep{Chase_LangChain_2022, schick2023toolformer}. 
\citet{zhang2023understanding} propose following this line to augment \acp{llm} 
with causal reasoning. 
However, similar to practitioners, \acp{llm} incorporating causal tools 
require a strategy for choosing the appropriate 
causal method for a particular problem.

We refer to \emph{causal method selection} as the task of
choosing the best method for a given causal task
from a candidate set $\{ M_1, M_2, \hdots, M_K \}$
for a given dataset $D_n$.
For example, if the causal task is structure learning,
then the methods would be various causal discovery algorithms.
Our goal in this work is to efficiently select the best method
for the given causal task from the input dataset.
The selection problem is especially challenging in the causal setting. 
Causal inference relies on strong assumptions.
Many such assumptions, like causal sufficiency, 
are inherently untestable~\citep{ashman2023causal, kong2023identification}. Even with domain knowledge justifying these assumptions,
multiple methods might be available.
Techniques like cross-validation do not apply
since evaluation depends on the true \ac{scm}.
However, many assumptions, such as linear or normality, are testable in principle~\citep{razali2011comparisions}.
Thus, observable features of a dataset can be 
used to guide the choice of the best causal method.
One can pick the most generally applicable method 
available, however, simpler methods might work better due to a limited sample size or fewer tuning parameters.
Moreover, beyond explicit assumptions, 
the performance of a causal method also depends on 
implicit dataset properties and optimization concerns.

In this work, inspired by foundation models \citep{bommasani2021opportunities}, 
we propose \ac{camp},
a framework for learning the best method for a given dataset
(Sec.~\ref{sec:causal-method-selection})
in a supervised manner.
Our work builds on prior supervised approaches for predicting the
causal structure from an input dataset \citep{ke2022learning, lorch2022amortized}.
We generate datasets from a diverse set of synthetic \acp{scm},
score all candidate methods on every dataset,
and then train a model to directly predict 
the best method for an input dataset.
Unlike traditional supervised learning, we make one prediction
per dataset and not per sample.
Given that knowledge of the underlying dataset
properties like linearity can aid method selection, 
we formulate a self-supervised pre-training
objective around predicting the dataset assumptions 
(Sec.~\ref{sec:cms-semi-supervised}),
and then fine-tune the pre-trained model on a 
limited labeled data.
The self-supervision provides a simple and effective way to 
inject useful inductive biases into the model,
improves computational efficiency,
requiring lesser labeled data as well as fewer training iterations.

Our approach allows learning 
implicit dataset features to predict which method 
is best for a dataset. This may include, but is not limited to 
testable assumptions made by the different methods.
At inference time, \ac{camp} can provide fast zero-shot 
predictions of the best method for a dataset.
Additionally, \ac{camp} can be applied to a wide range of causal tasks
such as causal discovery, treatment effect estimation, and covariate selection,
where the labels can be generated using synthetic \acp{scm}. 
In our experiments, we focus on the task of method selection for
causal discovery (Sec.~\ref{sec:experiments}).
We observe that, despite being trained on only synthetic data, 
\ac{camp} generalizes to synthetic 
out-of-distribution datasets and 
performs well on semi-synthetic and real-world 
gene expression benchmarks, outperforming simple heuristics for 
method selection like selecting the best method on average.

\section{Related Work}
\label{sec:related}
\paragraph{Non-Causal Model Selection} 
Our work is related to model selection which has been 
studied by many works in the context of supervised machine learning.
The common techniques are the
holdout validation method and
$k$-fold cross-validation, both of which use one partition of
the data for training and a different one for evaluation \citep{raschka2018model, arlot2010survey}.
The field of Automated machine learning (AutoML) aims to automate various parts of the
machine learning pipeline like feature and hyperparameter selection
aiming to reduce human effort in designing these pipelines
\citep{he2021automl, hutter2019automated}.
These methods do not directly apply in the causal setting since
evaluation requires knowledge of the true causal model, such as the true graph or counterfactual data,
which are not known for real problems.

\paragraph{Evaluating Causal Methods} 
Another line of work studies model evaluation and selection
for causal inference.
For conditional average treatment effect (CATE) estimation, many works have proposed data-driven
strategies for model selection based on estimates of the counterfactual risk 
\citep{rolling2014model, gutierrez17a, alaa2019validating, saito2020counterfactual}.
Empirical studies comparing numerous estimators and
model selection strategies for CATE estimation have also been conducted
revealing complex interactions between the underlying data assumptions, methods,
and selection strategies
\citep{schuler2018comparison, mahajan2022empirical, machlanski2023hyperparameter, curth2023search, matthieu2023select}.
For model selection and hyperparameter tuning in causal discovery,
various metrics have been used 
like the stability of the output across dataset perturbations
\citep{liu2010stability, raghu2018evaluation} and
multiple runs \citep{strobl2021automated},
the Bayesian Information Criterion \citep[Sec.~4.1]{maathuis2009est},
predictive performance of the learned causal model \citep{biza2020tuning, biza2022out},
and compatibility of the learned graphs across different 
subsets of variables \citep{faller2023self}.
In contrast, we take a supervised learning
approach to predict the best causal method and our strategy
can be viewed as complementary to these works.

\paragraph{Supervised Causal Inference} 
A closely related line of work frames causal discovery as a
supervised learning task. 
\citet{li2020supervised} study supervised causal discovery
using synthetic datasets for linear causal models,
predicting the causal graph from an input correlation matrix.
\citet{petersen2022causal} propose to learn an equivalence
class of causal graphs from an observational dataset
by training on simulated linear Gaussian data.
\citet{ke2022learning} propose a transformer-based model
that learns to map a dataset with observational and interventional samples
to the causal structure in a supervised manner using synthetic training data.
\citet{ke2023discogen} extend this work for discovering gene regularity networks.
\citet{wang2022meta} propose a supervised causal discovery method for
synthetic microprocessor and brain-network datasets.
\citet{lorch2022amortized} propose \emph{AVICI}, a variational 
inference model to predict the causal structure directly
from the input dataset, training on synthetically 
generated datasets.
We use the self-attention architecture of \emph{AVICI} to make
dataset-level predictions.
The architecture encodes desirable permutation invariances for 
improving the statistical efficiency of the predictor. 
While these works predict the causal structure in a supervised
manner, they do not consider method selection.

\section{Problem Formulation}
\label{sec:setup}
\begin{table}
\centering
\begin{tabular}{ ll } 
\toprule
\multicolumn{1}{c}{\textbf{Method}} & \multicolumn{1}{c}{\textbf{Description}} \\ 
\midrule
DirectLiNGAM \citep{shimizu2011directlingam} & DAG learning for linear non-Gaussian data \\
NOTEARS-linear \citep{zheng2018dags} & A gradient-based method for linear data \\
NOTEARS-MLP \citep{zheng2020learning} & MLP-based NOTEARS for nonlinear data \\
DAG-GNN \citep{yu2019dag} & DAG learning with Graph Neural Networks \\
GraNDAG \citep{lachapelle2019gradient} & DNN-based method for nonlinear additive noise data \\
DECI \citep{geffner2022deep} & Bayesian method for nonlinear additive noise data \\
\bottomrule
\end{tabular}
\caption{Brief description of the six candidate causal discovery methods.}
\label{table:candidate-causal-discovery-methods}
\end{table}

In this work, given a dataset $\X$, the aim is to choose the best causal method 
(as assessed by some score function)
from a candidate set $\{ M_1, M_2, \hdots, M_K \}$ at test time. 
The dataset is denoted as $\X \in \RB^{n \times V}$,
where $n$ is the number of samples and $V$ is the number of
variables (we assume real-valued data but our approach can
be generalized to discrete and mixed-type datasets).
There is a set of $K$ candidate methods: $\MM = \{ M_1, \hdots, M_K \}$
and a scoring function $s(M, \X; \theta_{*}) \in \RB$ 
for method $M \in \MM$ and dataset $\X$ that depends on some features of 
the true SCM (e.g., the true graph), 
denoted by $\theta_{*}$.
The goal is to develop a selection strategy $\SM : \RB^{n \times V} \mapsto \MM$ 
such that the selected method $\SM(\X)$ maximizes the score $s(\SM(\X), \X; \theta_{*})$. 
The score function determines how the methods are assessed
for the problem at hand. 
In the causal inference setting, this
score cannot be estimated using a holdout 
validation set since it depends on the unknown 
true causal model.

Throughout, we assume that the $n$ samples are generated i.i.d.
from (unknown) SCM associated with a 
directed acyclic graph (DAG) $G_*$ over $V$ nodes \citep[Sec.~6.2]{peters2017elements}.
Each node $X_j$, for $j \in [V]$, is generated according to the structural equation
$X_j := f_j( X_{\text{pa}(j; G_*)}, n_j )$, 
where $n_j$ is an exogenous noise term,
$\text{pa}(j; G_*)$ denotes the parents of $X_j$ in $G_*$,
and $f_i$ is an arbitrary function describing how $X_j$
depends on its parents and the noise term.

In this work, we focus on method selection for causal discovery
from observational data. 
In causal discovery, the aim is to discover the underlying causal
graph (or equivalence class thereof) from the dataset $\X$.
We consider six candidate causal discovery methods:
$\MM = \{$
DirectLiNGAM \citep{shimizu2011directlingam}, NOTEARS-linear \citep{zheng2018dags},
NOTEARS-MLP \citep{zheng2020learning}, DAG-GNN \citep{yu2019dag},
GraNDAG \citep{lachapelle2019gradient}, and DECI \citep{geffner2022deep} $\}$.
Every method $M \in \MM$ outputs an estimated DAG $\widehat{G}_M(\X)$.
These methods work with observational data and assume
causal sufficiency (i.e., no hidden variables), which also entails
that each noise term $n_j$ is independent of all other variables.
The candidate set $\MM$ was chosen to encompass methods for 
linear and nonlinear SCMs as well as recent 
gradient and neural network-based methods
(see Table~\ref{table:candidate-causal-discovery-methods} for a brief description of the methods).

We use the F1-score between the binary adjacency matrices of the true DAG
and the estimated DAG 
as the scoring function
for evaluating the causal discovery methods: 
$s(M, \X; G_*) = \text{F1}(\widehat{G}_M(\X), G_*)$.
The F1-score is a commonly used metric for evaluating the performance
of causal discovery algorithms and its range is agnostic to the size of the
graph (unlike structural hamming distance), enabling meaningfully
comparisons of the scores across different graph sizes.

\section{Learned Causal Method Prediction}
\label{sec:causal-method-selection}
\begin{figure}
\centering
\includegraphics[scale=0.80]{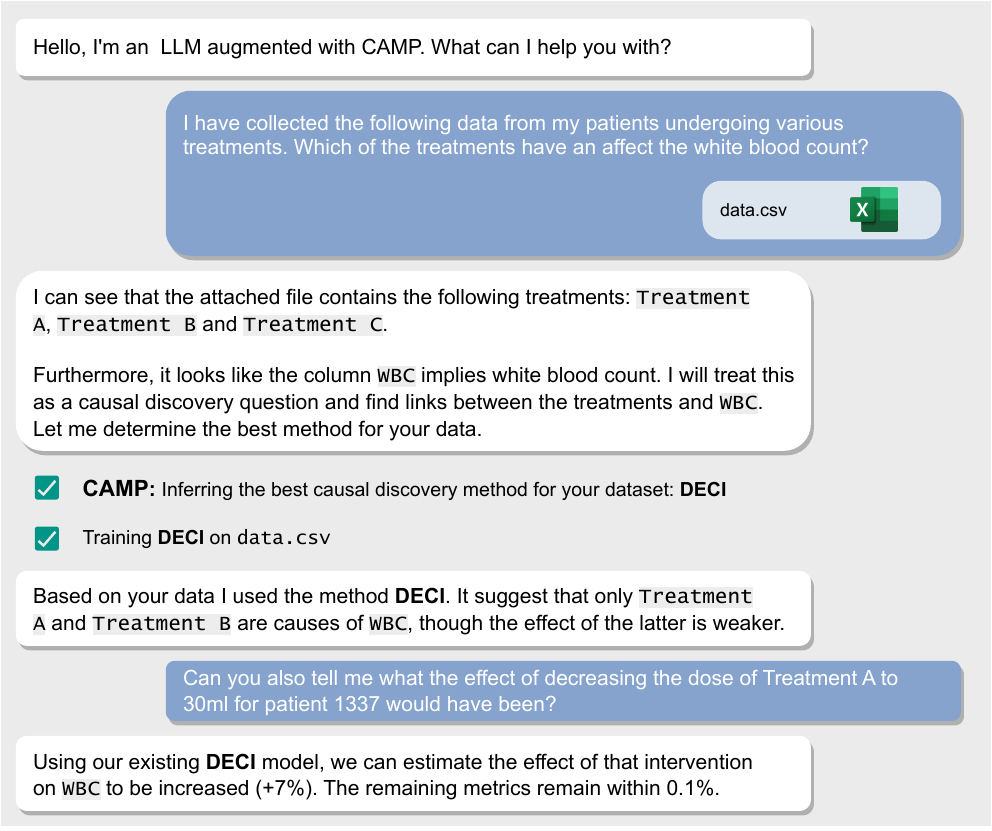}
\caption{
An illustration of an interaction with an \ac{llm} augmented with \emph{CAMP}. 
The LLM can parse the user's intent from their natural language query
and invoke \emph{CAMP} to answer causal questions,
allowing the practitioner to interact with causal methods effortlessly.
}
\label{fig:mockup}
\end{figure}

In this section, 
we describe CAusal Method Predictor (CAMP),
a framework for \emph{learning} the mapping from a dataset to
the best method. 
We describe two strategies for predicting the best method:
(i) Supervised (Sec.~\ref{sec:cms-supervised}):
we train a deep neural network (DNN) to directly classify the highest-scoring method from
the input dataset; and
(ii) Semi-supervised (Sec.~\ref{sec:cms-semi-supervised}): we propose a self-supervision strategy
to pre-train the DNN before fine-tuning it on labeled data,
improving the statistical and computational efficiency of the 
purely supervised approach.
The supervision allows CAMP to learn observable features
of the input dataset (beyond explicitly specified assumptions) 
that can be used to decide the best
causal method in a data-driven manner.

CAMP can be readily integrated with an \ac{llm} to form an augmented causal agent
that can answer causal questions from a given dataset 
(see Fig.~\ref{fig:mockup} for a demonstration of this interaction mode with
such an agent).
For a natural language query, the LLM can parse the user's intent
to determine the causal task they want to perform 
(e.g., causal discovery),
and invoke \emph{CAMP} as an external tool to determine the best causal 
method for that dataset and causal task.

\subsection{Supervised Causal Method Predictor (CAMP-Sup)} \label{sec:cms-supervised}

We treat causal method selection as
as a $K$-class classification task.
We generate labeled training data consisting of 
$(\X, L)$ instances, where
$\X$ is a dataset and $L \in \MM$ is the target candidate method,
and train a DNN to predict the target label $L$ from 
the input dataset $\X$.
Unlike traditional supervised learning, we make one prediction
\emph{per dataset} (not per sample).

\paragraph{Training data.}
We generate the training instances as follows:
(i) we sample a random SCM (see Sec.~\ref{sec:results-synthetic-data-gen} for details on the sampling) and generate a dataset $\X \in \RB^{n \times V}$ from that SCM with random $n$ and $V$;
(ii) we run each candidate method on the dataset $\X$ and score it
using the true graph $G_{*}$ (which is known since the SCM is synthetic); and
(iii) the target label $L$ is the method with the highest score: 
$L = \arg\max_{M \in \MM} s(M, \X)$.
The full training data is $(X_i,\, L_i)_{i=1}^T$.

\paragraph{Model architecture.}
We use an encoder-decoder-style model. 
The encoder takes the input dataset $\mathbf{X} \in \mathbb{R}^{n \times V}$
(with arbitrary $n$ and $V$)
and outputs a $Z$-dimensional embedding of the dataset.
We use the same encoder architecture as \citep[Sec. 4.2]{lorch2022amortized}.
The encoder is composed of $L$ identical layers. 
The crux of each layer is an alternating multi-headed self-attention:
the first self-attention attends across the $V$ axis, treating $n$ as the batch dimension;
the second self-attention attends across the $n$ axis, treating $V$ as the batch dimension.
After $L$ such layers, the output dimension is $(n \times V \times Z)$,
where $Z$ is the embedding dimension of the muli-headed self-attention.
The attention layers allow the network to aggregate information across both the
sample and node axes as well as process an arbitrary-sized dataset.
Similar to \citet{lorch2022amortized}, we then apply a max-pooling across the $n$ and $V$ axes, 
resulting in a $Z$-dimensional dataset embedding.
This embedding is permutation invariant across both the $n$ and $V$ axes:
this is desirable because the prediction should be agnostic to the order
of the samples (since they are i.i.d.) and the nodes.
The decoder is a fully-connected feedforward network (FFN) with 
a $K$-dimensional output representing the classification logits
for each of the candidate methods.
Although we use the same DNN architecture as \citet{lorch2022amortized},
our goal is to predict the best method. In contrast, they attempt to
learn the DAG directly in a supervised manner and in principle,
their method \emph{AVICI} can be added to the list of candidate methods that
we select among.

\paragraph{Loss function.}
We treat method selection as a multi-class classification problem
and train the DNN end-to-end with the cross-entropy loss:
\begin{equation}
L_{\text{CE}}(p, L) = - \sum_{i=1}^{K} L_i \log (p_i),   
\end{equation}
where $p \in \Delta^{K - 1}$ are the predicted probabilities and $L \in \{0, 1 \}^{K}$
is the one-hot vector denoting the target label.
Since we also have access to the raw scores of all candidate methods for
each dataset, rank-based loss functions can also be used \citep{wang2018lambdaloss}.
Empirically, we did not see an improvement with these loss functions
and so we used the cross-entropy loss in our experiments 
(see Fig.~\ref{fig:apdx-rank-losses} in Appendix~\ref{apdx:rank-based-losses}).

\begin{figure}[t]
\centering
\includegraphics[scale=0.54]{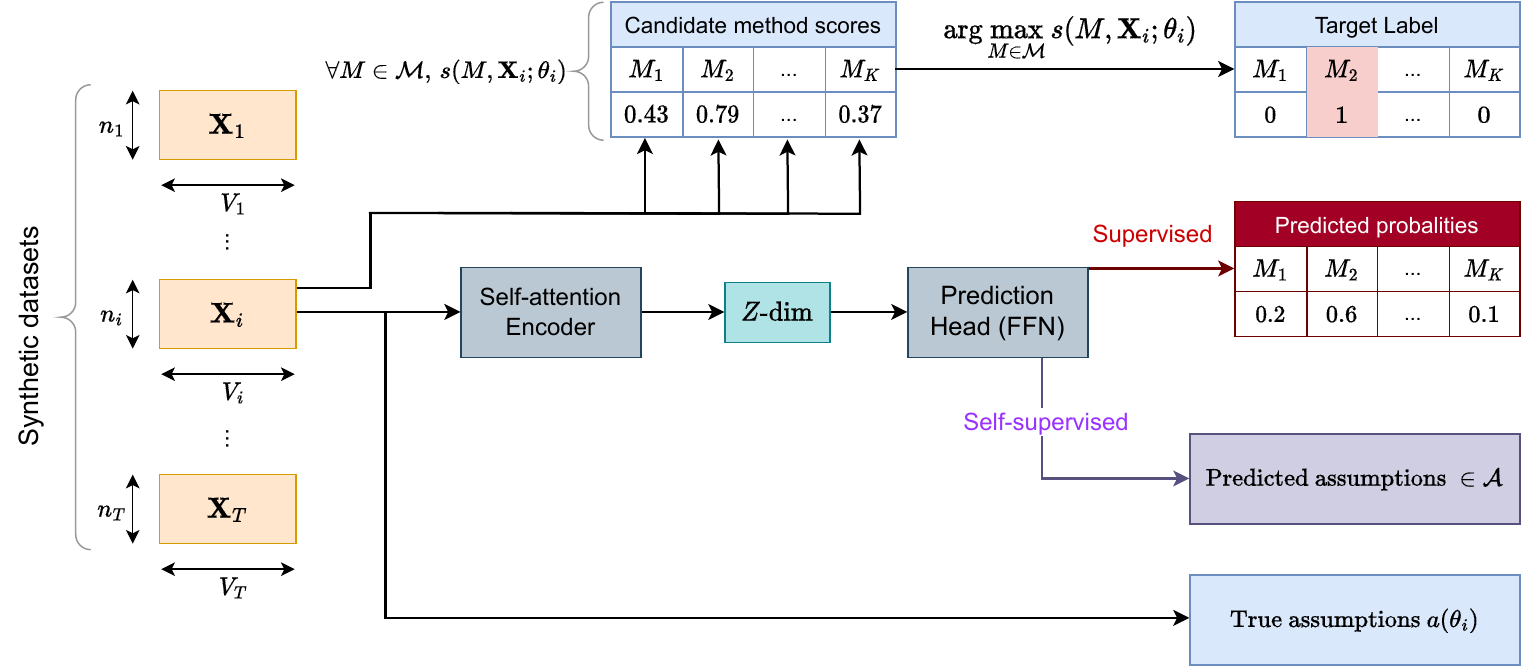}
\caption{Training pipelines for \emph{CAMP-Sup} and \emph{CAMP-SemiSup}.
For \emph{CAMP-Sup}, we generate synthetic datasets, score each candidate, and classify 
the highest-scoring method.
For \emph{CAMP-SemiSup}, there is a self-supervised pre-training step 
to predict the SCM assumptions, followed by supervised fine-tuning 
on a limited amount of labeled data.}
\label{fig:training-pipeline}
\end{figure}

\paragraph{Inference.}
At inference time, we perform a forward pass through the trained DNN
to get the best method for the input dataset.
The supervised approach enables zero-shot prediction of 
the best method,
which is fast, because we do not
need to run all candidate methods at inference time.
The model learns a mapping from implicit dataset features 
to the best method in a data-driven manner.

\subsection{Semi-Supervised Causal Method Predictor (CAMP-SemiSup)} \label{sec:cms-semi-supervised}

Generating labels can be computationally expensive 
because
it requires running all candidate methods on every dataset.
This is especially problematic in the case of causal discovery
where running even a single method can take a long time.
Addressing this concern, we propose a
semi-supervised approach based on 
self-supervised pre-training followed by supervised fine-tuning 
\citep{zhai2019s4l, chen2020big}.

Inferring causality from observational data always requires some assumptions
on the data-generating process.
In causal discovery, different methods 
utilize different sets of assumptions on the SCM 
and knowledge of these assumptions can be informative 
in determining the best method.
So we formulate a self-supervised pre-training objective 
around predicting the assumptions that hold in the 
underlying causal model that are relevant to the causal
task at hand (e.g., whether linearity or normality holds). 
This injects useful inductive biases into 
the DNN prior to the supervised finetuning step.

Let $\Theta$ denote some set of SCMs and 
$\AM$ denote some set of (testable) 
assumptions of an SCM. 
Consider a function $a : \Theta \mapsto \AM$ that
maps an SCM to the assumptions that hold for that SCM.
The set $\AM$ can represent any assumptions that might be 
useful for selecting among the candidate methods.
For synthetically constructed SCM and dataset pairs $(\theta_i, \X_i)_{i=1}^{T}$,
where $\theta_i \in \Theta$ and $\X_i$ is generated from SCM $\theta_i$, 
we train an encoder-decoder DNN similar to Sec.~\ref{sec:cms-supervised} 
to predict $a(\theta_i)$ from the input dataset $\X_i$.
Generating training data for this self-supervised task is relatively 
cheap: $a(\theta_i)$ can directly be 
determined from the synthetic SCM $\theta_i$ 
without running any of the candidate methods.
After training this DNN, we take this pre-trained encoder
and fine-tune it on limited labeled data as described in Sec.~\ref{sec:cms-supervised}.
For our experiments, we use $\AM = \{ \text{Linear Gaussian}, \text{Linear non-Gaussian}, \text{Nonlinear} \}$. 
Since $\AM$ is a discrete set in our experiments, we
treat it as a $3$-way classification task and
use the cross-entropy loss for pre-training.

In practice, the space $\AM$ can contain other properties of
an SCM that can be useful for determining the best causal method, 
like the presence and type of identifiable unmeasured confounders
(e.g., bow-free confounding \citep{ashman2023causal}), 
characteristics of the DAG (e.g., sparsity), etc.
The set $\AM$ should be chosen depending
on the causal task as well as the candidate methods
(\citet{berrevoets2023causal} discuss several axes for 
grouping assumptions for causal inference).
In our experiments, we show that the semi-supervised model
not only requires lesser labeled data and training steps,
but also generalizes better beyond the training distribution.
Thus, semi-supervision 
can also be useful for scaling up the supervised approach to a 
larger set of methods.

\begin{remark}
One strategy for method selection is to leverage the prediction of
the set $a(\theta)$ by mapping each item in $\AM$ to a method.
Our approach is more flexible since we learn this mapping in a 
data-driven way. 
Moreover, during supervision, our approach can also learn to use dataset features 
that might be difficult to explicitly elicit in the space $\AM$.
\end{remark}

\section{Experimental Results}
\label{sec:experiments}
We evaluate our causal method selection strategy 
for synthetic datasets (Sec.~\ref{sec:results-synthetic-data})---both in-distribution and 
out-of-distribution---as well as four semi-synthetic and real-world
benchmark datasets (Sec.~\ref{sec:results-real-world}).

\subsection{Synthetic training data generation} \label{sec:results-synthetic-data-gen}

\paragraph{SCM and dataset generation.} 
We generate datasets with varying sample and graph sizes from 
a diverse set of linear and nonlinear SCMs.
We consider sample sizes $n \in [600, 1200]$ and graph sizes
$V \in [8, 12]$.
The graph is sampled from an Erdos-Renyi distribution
with edge probabilities uniformly sampled from $[0.3, 0.7]$.
We consider a diverse set of SCMs that encompass several
linear and nonlinear SCMs considered in the causal discovery literature.
Our training data contains datasets from the following causal models:
(1) \emph{Linear Gaussian}: We simulate $X_j = w^\top_j X_{\text{pa}(j)} + n_j$,
where the coefficients $w_j$ are uniformly random and
$n_j$ is Gaussian (similar to \citet{zheng2018dags});
(2) \emph{Linear non-Gaussian}: The same as \emph{}{Linear Gaussian}, but
with $n_j$ belonging to a uniform or exponential distribution;
(3) \emph{Nonlinear Additive Noise Models (ANM)}: We simulate $X_j = f_j(X_{\text{pa}(j)}) + n_j$, where $n_j$ is Gaussian and each $f_j$ is one of two nonlinear functions 
(similar to \citet{zheng2020learning}):
(i) random function from a Gaussian Process (GP), or
(ii) (Additive GP) $f_i(X_{\text{Pa}(i)}) = \sum_{j \in \text{Pa}(i))} g_j(X_j)$, where
each $g_j$ is a random function from a GP;
(4) \emph{Post-nonlinear (PNL) model} \citep{zhang2012identifiability}:
We simulate $X_j = f_j(g_j(X_{\text{pa}(j)} + n_j))$ where $g_i, f_i$
are nonlinear functions from one of the following PNL models:
(i) $f_i$ and $g_i$ are sampled as weighted sums of GPs and sigmoids \citep{uemura2022multivariate}, or
(ii) $f_i$ is a polynomial and $g_i$ is the cube-root \citep{keropyan2023rank};
(5) \emph{Location-scale model} \citep{immer2023identifiability}: This is a heteroskedastic noise model with
$X_j = f_j(X_{\text{pa}(j)}) + g_j(X_{\text{pa}(j)}) \cdot n_j$,
where $n_j \sim \NM(0, \sigma^2_j)$ and $f_j, g_j$ are random functions from a GP.
The training distribution contains an $\approx 11$--$23$--$66\%$ split of linear non-Gaussian,
linear Gaussian, and nonlinear SCMs, respectively;
and amongst nonlinear SCMs, we generate each type with equal probability
(see Appendix~\ref{apdx:synthetic-data-details} for additional details on the synthetic data distribution).

\paragraph{Target label.}
We consider six causal discovery methods
(see Table~\ref{table:candidate-causal-discovery-methods} and Sec.~\ref{sec:setup}).
For all methods except DECI,
we use the implementations from the library \emph{gCastle} \citep{zhang2021gcastle}
(we use the default hyperparameters from gCastle for all methods).
For DECI, we use the Gaussian noise implementation from the 
\emph{causica} package \citep{kiciman2022causal}.
For each synthetically generated dataset, 
we run the six causal discovery methods and compute the F1-score
using the true DAG. Since DECI is a Bayesian method, 
we compute the average F1-score across $1000$ draws from its posterior
distribution over DAGs.
The target label is a one-hot vector denoting the method with the highest F1-score.

\begin{figure}
\centering
\includegraphics[scale=0.40]{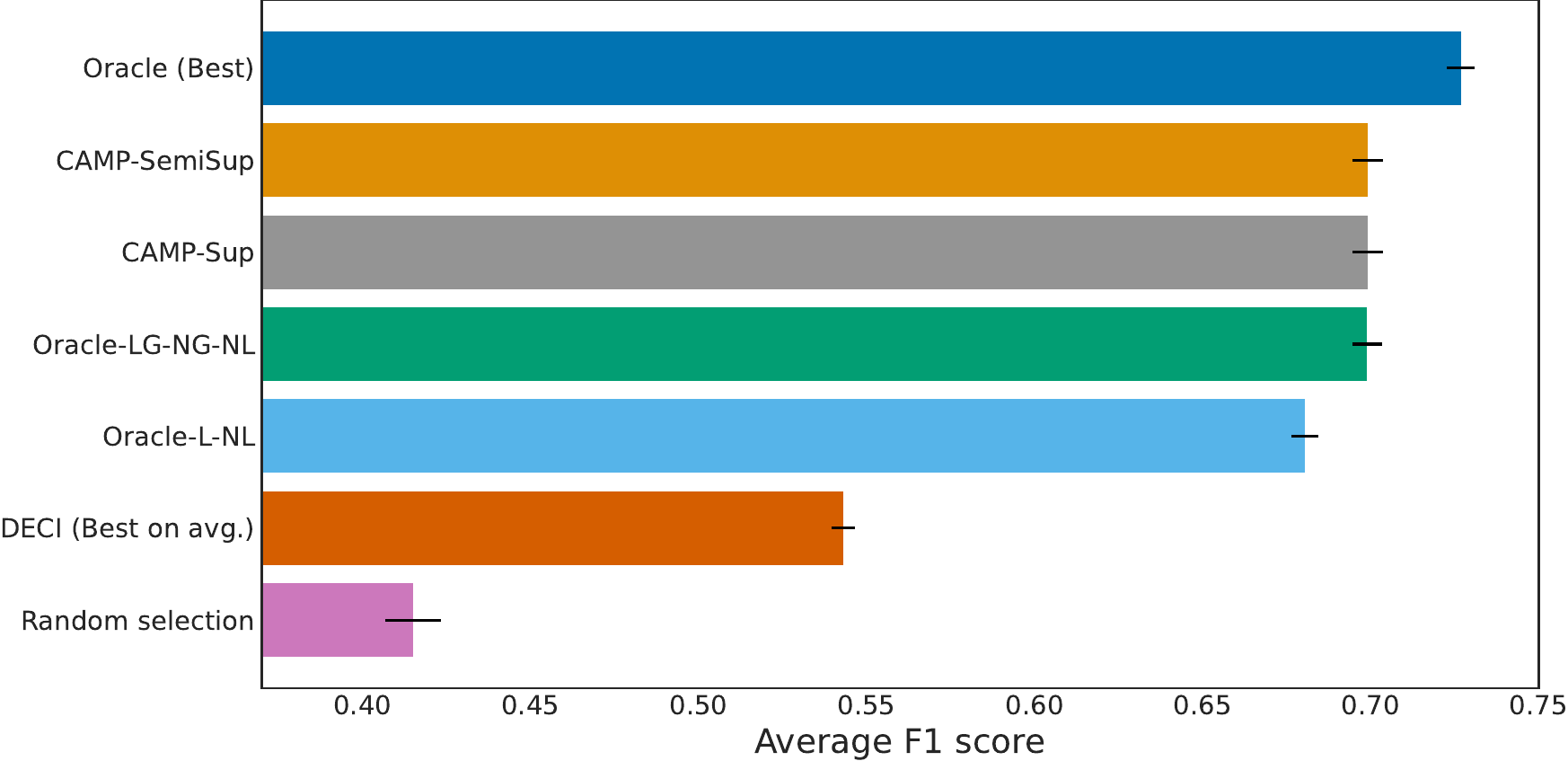}
\caption{Comparison of various strategies for selecting among six causal discovery methods (Table~\ref{table:candidate-causal-discovery-methods}) across a diverse set of SCMs (error bars denote 95\% CIs). 
\emph{Oracle (Best)} chooses the best method for each dataset.
The \emph{Oracle-LG-NG-NL} and \emph{Oracle-L-NL} oracles select the best method depending 
on the dataset type, improving over
\emph{DECI} (the best method on average).
Both \emph{CAMP} methods match the score of \emph{Oracle-LG-NG-NL}.
}
\label{fig:comparison-of-cd-methods}
\end{figure}

\subsection{Results on synthetic data} \label{sec:results-synthetic-data}

We first evaluate the supervised and semi-supervised approaches on a test set
with the same distribution as the training set (as described in Sec.~\ref{sec:results-synthetic-data-gen}).
For the results in this section, 
we used validation and test sets of $2000$ and $3414$ datasets, respectively.
For the encoder, we use $L = 4$ layers and attention embedding size $Z = 32$;
and for the decoder, we used a FFN with $2$ hidden layers of size $32$.
For the semi-supervised approach, we used $\approx 50,000$ datasets
for the self-supervised pre-training step, and 
we train the model to predict the underlying SCM assumptions
from the set $\AM = \{ \text{Linear Gaussian}, \text{Linear non-Gaussian}, \text{Nonlinear}\}$.
The model achieved a nearly perfect accuracy for this task after the 
pre-training step.

\begin{figure}
\centering
\subfigure[Labeled data requirements]{
\includegraphics[scale=0.47]{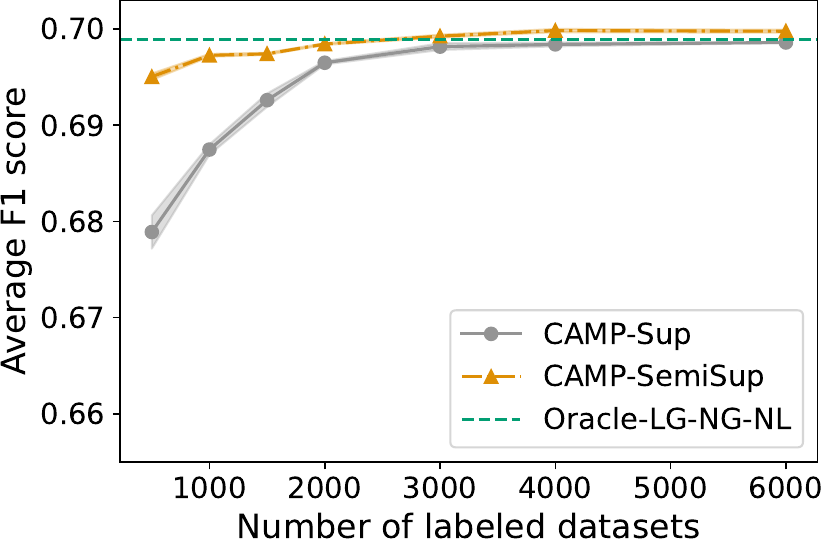}
\label{fig:results-iod-semi-sup-data-size}}
\hspace{20pt}
\subfigure[Rate of convergence]{
\includegraphics[scale=0.47]{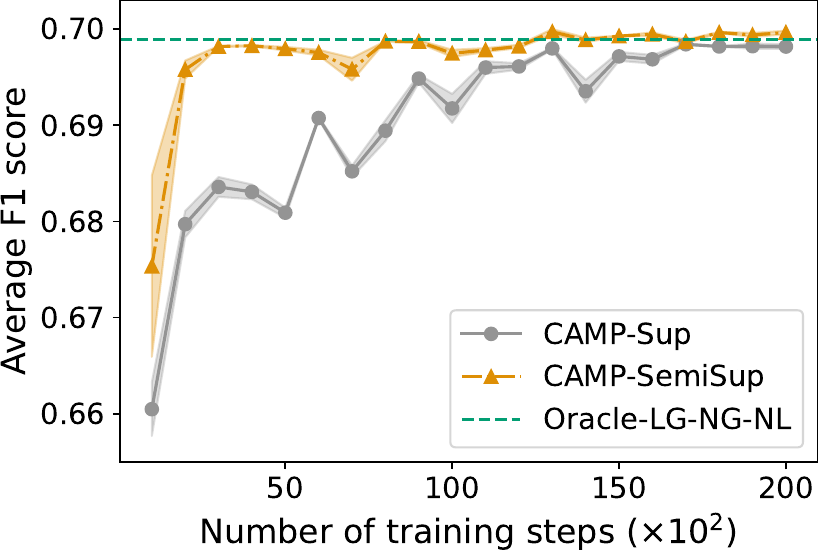}
\label{fig:results-iod-semi-sup-train-steps}}
\caption{Average F1 scores of \emph{CAMP-Sup} and \emph{CAMP-SemiSup}
on the test set
(errors bars denote standard errors over $4$ runs).
\emph{CAMP-SemiSup} requires 
(a) lesser training labeled data than \emph{CAMP-Sup} 
(with $3000$ labels, \emph{CAMP-SemiSup} outperforms \emph{CAMP-Sup} with
$6000$ labels); and
(b) fewer training steps to converge.
For \emph{CAMP-SemiSup}, $\approx 50,000$ datasets were used for pre-training.}
\label{fig:results-in-distribution}
\end{figure}

We show the performance of 
our supervised (\emph{CAMP-Sup}) and 
semi-supervised (\emph{CAMP-SemiSup}) models, 
trained on $6000$ labeled datasets (see Fig.~\ref{fig:comparison-of-cd-methods}),
a test set with the same distribution as the training set.
We also plot the following oracles to select the best method:
\emph{Oracle (Best)} selects the best method for every
dataset;
\emph{Oracle-LG-NG-NL} selects the 
best method based on whether the
SCM is linear Gaussian, linear non-Gaussian, or nonlinear;
and \emph{Oracle-L-NL} based on whether the
SCM is linear or nonlinear
(see Fig.~\ref{fig:apdx-synthetic-data-scores} in Appendix~\ref{apdx:synthetic-data-details}
for the average F1 scores of the six methods across different SCM types).
There is a substantial gap 
between \emph{Oracle (Best)}
and always picking the best method on average (DECI).
As the oracles leverage more knowledge of the dataset properties, their average scores increase relative to \emph{DECI} with \emph{Oracle-LG-NG-NL} outperforming
\emph{Oracle-L-NL}.
We see that \emph{CAMP-Sup} and \emph{CAMP-SemiSup} perform comparably.
Both \emph{CAMP} models significantly outperform always selecting DECI, 
the best method on average in the training set,
as well as the randomly selecting a candidate method 
(Fig.~\ref{fig:comparison-of-cd-methods}).
Next, we observe that \emph{CAMP} matches the performance of the oracle
\emph{Oracle-LG-NG-NL}.
This demonstrates the ability of \emph{CAMP} to automatically 
learn the mapping from implicit dataset features 
to the best method.

We also compare the average scores of \emph{CAMP-Sup} and 
\emph{CAMP-SemiSup} on the test set 
across different amounts of labeled training data 
(see Fig.~\ref{fig:results-iod-semi-sup-data-size}).
We observe that both approaches converge to \emph{Oracle-LG-NG-NL}
as the number of training data points gets large. 
But we see that the \emph{CAMP-SemiSup} achieves the same average score 
with fewer labeled data points:
$3000$ training datasets suffice for \emph{CAMP-SemiSup} to match the score
of \emph{CAMP-Sup} with $6000$ labeled datasets.
Moreover, the differences in the scores are much larger
when the number of labeled data points is small ($< 2000$),
showing the advantages of the semi-supervised approach in
small data regimes.
Next, we compare \emph{CAMP-Sup} and \emph{CAMP-SemiSup} in
terms of how quickly they converge (see Fig.~\ref{fig:results-iod-semi-sup-train-steps}).
We fix the amount of labeled data points to $6000$ and
compare the average scores of the two approaches on the test set
across increasing training steps.
Although both approaches converge as we train for long enough,
\emph{CAMP-SemiSup} requires significantly fewer training steps:
after $\approx 3000$ training steps, it achieves the same score as
the \emph{CAMP-Sup} after $\approx 12000$ steps.
These results show the computational and statistical advantages of the 
semi-supervised approach with a pre-training objective around the SCM
assumptions.

\begin{figure}
\centering
\includegraphics[scale=0.40]{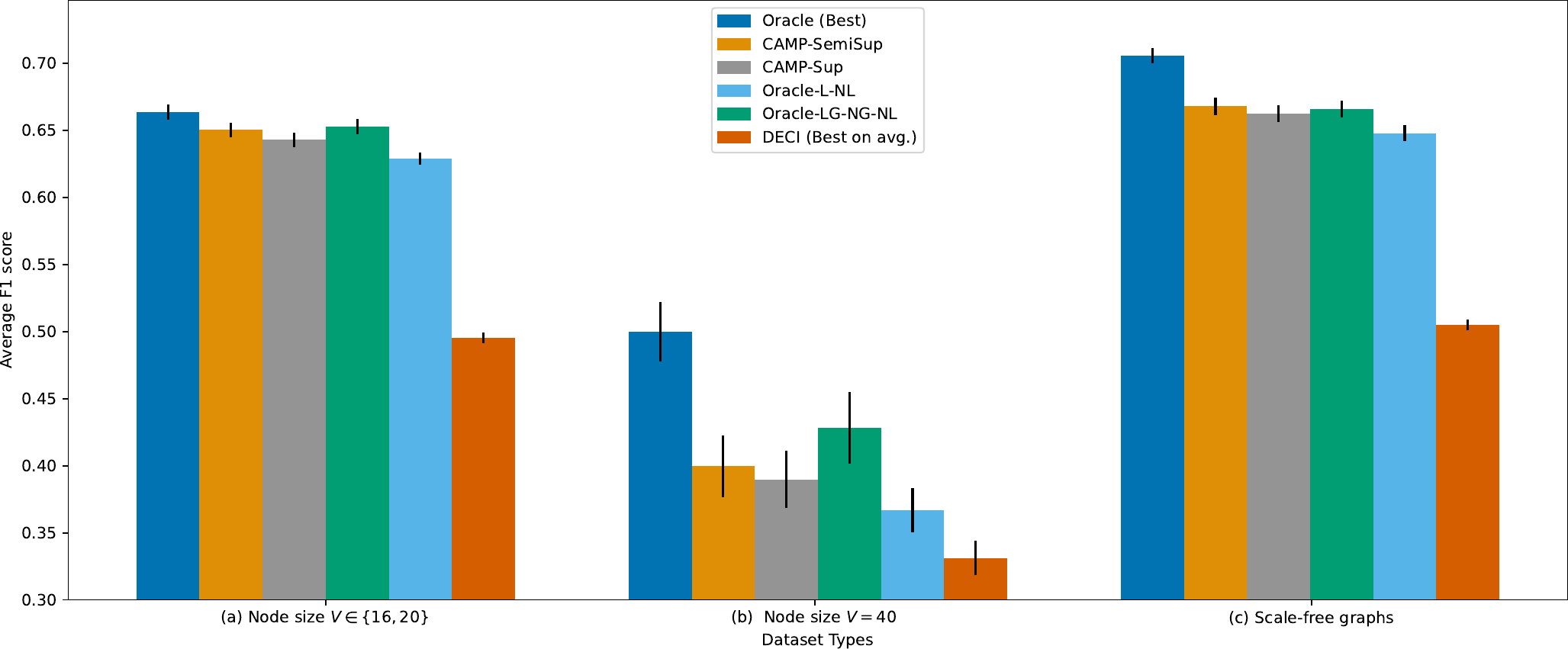}
\caption{Results on synthetic OOD datasets for 
(a), (b): graphs with a larger number nodes and 
(c): scale-free graphs.
\emph{CAMP} continues to perform well in regimes (a) and (c).
For (b), \emph{CAMP} still outperforms \emph{DECI} and
\emph{Oracle-L-NL}.
}
\label{fig:results-ood}
\end{figure}

\paragraph{Out-of-distribution datasets.}
We also test the generalization abilities of our method selection approach
beyond the training distribution (see Fig.~\ref{fig:results-ood}).
We evaluate our methods on synthetic SCMs with a larger number of nodes than 
encountered during training. 
When we increase the number of nodes to $V \in \{16, 20\}$ (Fig.~\ref{fig:results-ood}a),
the model continues to perform well.
Further increasing the node size to $V = 40$ worsens
the performance relative to \emph{Oracle (Best)} but is still
better \emph{Oracle-L-NL},
with both \emph{CAMP} models performing worse than the 
\emph{Oracle-LG-NG-NL} oracle, showing that much larger graphs can be
a challenge for \emph{CAMP} (Fig.~\ref{fig:results-ood}b).
Next, we test our methods on synthetic datasets generated for 
scale-free graphs \citep{barabasi1999emergence} (Fig.~\ref{fig:results-ood}c),
showing that the method continues to generalize to a different graph distribution.
In all three cases, both \emph{CAMP} models significantly outperform
\emph{DECI} (the best method on average on the training set).

\subsection{Results on semi-synthetic and real-world data} \label{sec:results-real-world}

An important concern with supervised training on only synthetic SCMs
is the extent to which the selection algorithm generalizes to
real-world datasets.
Addressing this concern, we demonstrate promising results
on the four semi-synthetic and real-world gene expression benchmarks.

We consider the following benchmarks:
(i) semi-synthetic datasets generated 
from the graphs \emph{MAGIC-NIAB} and \emph{MAGIC-IRRI} from
\emph{bnlearn} \citep{scutari2010learn, scutari2014multiple};
(ii) the \emph{SynTReN} generator \citep{van2006syntren} that creates synthetic 
gene regulatory networks and simulates gene expression data that approximates
experimental data; and
the real-world \emph{protein cells} dataset from \citet{sachs2005causal} 
\footnote{We use the dataset from {\small \url{https://github.com/cmu-phil/example-causal-datasets/}}.} 
which is commonly used to benchmark causal discovery algorithms.
The  \emph{MAGIC-NIAB} and \emph{MAGIC-IRRI} are linear Gaussian
SCMs with graph sizes $V = 44$ and $V = 64$, respectively.
For both graphs, we sample datasets with $n = 1000$ samples,
and use the SCM parameters from \emph{bnlearn}.
Both these graphs are significantly larger than the training distribution
of CAMP.
For \emph{SynTReN}, we simulate a dataset with $V = 20$ nodes and 
$n = 400$ samples.
The \citet{sachs2005causal} protein cells dataset contains $V = 11$
nodes and $n = 7466$ samples.

For each of the four benchmark datasets, we first run the six
candidate causal discovery methods $10$ times and compute the average
F1 score of each method 
(see Fig.~\ref{fig:apdx-results-real-world} in Appendix~\ref{apdx:real-world}
for the F1 scores of the six methods on each benchmark).
We compare the true ranking of the six methods on each benchmark
against the rankings predicted by \emph{CAMP} 
(see Table~\ref{tab:results-real-world}; the complete rankings are in 
Table~\ref{tab:apdx-real-world-rankings} in Appendix~\ref{apdx:real-world}). 
We first compare the scores of the top-ranked methods from \emph{CAMP}
against the true scoring method (\emph{Oracle Rank $1$}) on each dataset 
(see Table~\ref{tab:results-real-world}a).
We observe that \emph{CAMP-SemiSup} predicts the best method for all benchmarks
except \emph{MAGIC-IRRI}.
In contrast, the best-predicted method from \emph{CAMP-Sup} 
performs poorly on the \emph{bnlearn} graphs 
(the scores are close to zero because it picks a method that returns a nearly empty graph), but it successfully
picks out the best method for \emph{SynTReN} and the \emph{protein cells} benchmarks.

\begin{table}
\centering
\begin{tabular}{lcccc}
\toprule
 &  \textbf{MAGIC-NIAB} &  \textbf{MAGIC-IRRI} &  \textbf{SynTReN} & \textbf{Protein cells} \\
\midrule
\multicolumn{5}{l}{\textbf{a) Top-1 F1-score}} 
\vspace{5pt}
\\
Oracle & $0.28$ & $0.36$  & $0.25$  & $0.48$ \\
CAMP-SemiSup & $\textbf{0.28}$  & $0.31$ & $\textbf{0.25}$ & $\textbf{0.48}$ \\
CAMP-Sup & $0.00$ & $0.02$ & $\textbf{0.25}$  & $\textbf{0.48}$ \\
\midrule
\multicolumn{5}{l}{\textbf{b) Weighted F1-score}} 
\vspace{5pt}
\\
CAMP-SemiSup & $\textbf{0.26}$ & $\textbf{0.32}$  & $0.22$ & $0.35$ \\
CAMP-Sup & $0.07$ & $0.15$ & $\textbf{0.23}$ & $\textbf{0.38}$ \\
Average score & $0.18$ & $0.23$ & $0.16$ & $0.32$ \\
\midrule
\multicolumn{5}{l}{\textbf{c) Spearman's correlation coefficient}} 
\vspace{5pt}
\\
CAMP-SemiSup & $\textbf{0.60}$ & $\textbf{0.42}$ & $0.60$  & $\textbf{0.54}$ \\
CAMP-Sup & $-0.37$ & $-0.77$ & $\textbf{0.71}$  & $0.31$ \\
\bottomrule
\end{tabular}
\caption{F1 scores of the predicted top-1 method, the average of all methods weighed by their predicted probabilities, and the Spearman correlation between the true and predicted ranks.
We see that \emph{CAMP-SemiSup} performs well across all four benchmarks.
}
\label{tab:results-real-world}
\end{table}

Both \emph{CAMP-Sup} and \emph{CAMP-SemiSup} output a vector $p \in \Delta^{K-1}$ (probability simplex),
denoting the predicted probability of each method.
We compare the average score weighted by this probability:
$\sum_{i=1}^{K} p_i s(M_i, \X)$,
to the average score with equal weight given to each method
(\emph{Average score}):
$\sum_{i=1}^{K} \frac{1}{|\MM|} s(M_i, \X)$
(see Table~\ref{tab:results-real-world}b).
We observe that \emph{CAMP-SemiSup} always outperforms the \emph{Average score}
baseline for all benchmarks,
and \emph{CAMP-Sup} is worse on the \emph{bnlearn} graphs but significantly
better
for both \emph{SynTReN} and the \emph{protein cells} dataset.

Next, we present the Spearman's correlation coefficient, a measure
of agreement between the true ranking of
the methods and the ranking determined by their predicted probabilities
(see Table~\ref{tab:results-real-world}c).
Again, we see that the rankings of \emph{CAMP-SemiSup} are always
positively correlated whereas \emph{CAMP-Sup} leads to poor results
on the \emph{bnlearn} graphs.

Overall, these results show that \emph{CAMP-SemiSup} performs well on
these benchmarks and selecting the best predicted method leads to a
good F1-score.
This shows that \emph{CAMP} may generalize to unseen real-world
distributions despite having been trained only on synthetic data.
As for the good scores of \emph{CAMP-SemiSup} on the \emph{bnlearn}
graphs (where \emph{CAMP-Sup} fails), we conjecture that it is due to
the pre-trained inductive biases: 
the assumption predictor trained during the pre-training step correctly
predicts that both \emph{bnlearn} graphs are linear Gaussian, which
potentially allows \emph{CAMP-SemiSup} to perform much better than
\emph{CAMP-Sup}.
This demonstrates that semi-supervision leads to better generalization
in practice than the purely supervised approach.

\section{Discussion}
\label{sec:conclusion}
We leverage large-scale predictive models to select the best causal
method by framing the selection problem as a 
(semi) supervised learning task.
We generate datasets from a diverse set of synthetic SCMs 
and train a model to directly predict the best method from
the input dataset.
Using synthetic SCMs allows us to construct a large
amount of labeled training data. 
Our experimental results show
that CAMP performs favorably against several oracles 
and shows promising generalization beyond
the training distribution on common causal discovery benchmark datasets.
Moreover, CAMP can be integrated into an
an \ac{llm} toolchain to allow users to ask causal questions
about their datasets. 

Although our experiments focus on method selection for 
causal discovery, our proposed strategy is more generally applicable. 
In future work, we hope to test our selection strategy 
on other causal tasks like treatment effect estimation and covariate 
selection.
Moreover, theoretically understanding the generalization abilities and
limits of the supervised approach for causal tasks is also a promising
future direction.
Beyond causality, we hope to inspire other areas that 
may benefit from data-driven supervised method selection. 
For example, one such area is ODE solvers, 
where factors like problem stiffness may determine which 
method performs best~\citep{postawa2020comprehensive, dallas2017comparison}.
More broadly, our approach can be applied to tasks where there 
are several candidate methods, and synthetically generated datasets
can be used to generate labeled data for learning 
the mapping from a dataset to the best method.

\section*{Acknowledgements}

We thank Chao Ma for 
insightful comments that improved this work.
We thank members of the Causica team at 
Microsoft Research for helpful discussions.
We thank Colleen Tyler, Maria Defante, and Lisa Parks for conversations on real-world use
cases that inspired this work.

\bibliographystyle{abbrvnat}
\bibliography{refs}

\begin{thebibliography}{69}
\providecommand{\natexlab}[1]{#1}
\providecommand{\url}[1]{\texttt{#1}}
\expandafter\ifx\csname urlstyle\endcsname\relax
  \providecommand{\doi}[1]{doi: #1}\else
  \providecommand{\doi}{doi: \begingroup \urlstyle{rm}\Url}\fi

\bibitem[Alaa and Van Der~Schaar(2019)]{alaa2019validating}
A.~Alaa and M.~Van Der~Schaar.
\newblock Validating causal inference models via influence functions.
\newblock In \emph{International Conference on Machine Learning}, 2019.

\bibitem[Arlot and Celisse(2010)]{arlot2010survey}
S.~Arlot and A.~Celisse.
\newblock {A survey of cross-validation procedures for model selection}.
\newblock \emph{Statistics Surveys}, 2010.
\newblock \doi{10.1214/09-SS054}.

\bibitem[Ashman et~al.(2023)Ashman, Ma, Hilmkil, Jennings, and Zhang]{ashman2023causal}
M.~Ashman, C.~Ma, A.~Hilmkil, J.~Jennings, and C.~Zhang.
\newblock Causal reasoning in the presence of latent confounders via neural {ADMG} learning.
\newblock In \emph{International Conference on Learning Representations}, 2023.

\bibitem[Barab{\'a}si and Albert(1999)]{barabasi1999emergence}
A.-L. Barab{\'a}si and R.~Albert.
\newblock Emergence of scaling in random networks.
\newblock \emph{Science}, 1999.

\bibitem[Berrevoets et~al.(2023)Berrevoets, Kacprzyk, Qian, and van~der Schaar]{berrevoets2023causal}
J.~Berrevoets, K.~Kacprzyk, Z.~Qian, and M.~van~der Schaar.
\newblock Causal deep learning.
\newblock \emph{arXiv preprint arXiv:2303.02186}, 2023.

\bibitem[Biza et~al.(2020)Biza, Tsamardinos, and Triantafillou]{biza2020tuning}
K.~Biza, I.~Tsamardinos, and S.~Triantafillou.
\newblock Tuning causal discovery algorithms.
\newblock In \emph{International Conference on Probabilistic Graphical Models}, 2020.

\bibitem[Biza et~al.(2022)Biza, Tsamardinos, and Triantafillou]{biza2022out}
K.~Biza, I.~Tsamardinos, and S.~Triantafillou.
\newblock Out-of-sample tuning for causal discovery.
\newblock \emph{IEEE Transactions on Neural Networks and Learning Systems}, 2022.

\bibitem[Bommasani et~al.(2021)Bommasani, Hudson, Adeli, Altman, Arora, von Arx, Bernstein, Bohg, Bosselut, Brunskill, et~al.]{bommasani2021opportunities}
R.~Bommasani, D.~A. Hudson, E.~Adeli, R.~Altman, S.~Arora, S.~von Arx, M.~S. Bernstein, J.~Bohg, A.~Bosselut, E.~Brunskill, et~al.
\newblock On the opportunities and risks of foundation models.
\newblock \emph{arXiv preprint arXiv:2108.07258}, 2021.

\bibitem[Bubeck et~al.(2023)Bubeck, Chandrasekaran, Eldan, Gehrke, Horvitz, Kamar, Lee, Lee, Li, Lundberg, et~al.]{bubeck2023sparks}
S.~Bubeck, V.~Chandrasekaran, R.~Eldan, J.~Gehrke, E.~Horvitz, E.~Kamar, P.~Lee, Y.~T. Lee, Y.~Li, S.~Lundberg, et~al.
\newblock Sparks of artificial general intelligence: Early experiments with gpt-4.
\newblock \emph{arXiv preprint arXiv:2303.12712}, 2023.

\bibitem[Burges et~al.(2005)Burges, Shaked, Renshaw, Lazier, Deeds, Hamilton, and Hullender]{burges2005learning}
C.~Burges, T.~Shaked, E.~Renshaw, A.~Lazier, M.~Deeds, N.~Hamilton, and G.~Hullender.
\newblock Learning to rank using gradient descent.
\newblock In \emph{International Conference on Machine learning}, 2005.

\bibitem[Burges(2010)]{burges2010ranknet}
C.~J. Burges.
\newblock From ranknet to lambdarank to lambdamart: An overview.
\newblock \emph{Learning}, 2010.

\bibitem[Chase(2022)]{Chase_LangChain_2022}
H.~Chase.
\newblock {LangChain}, Oct. 2022.
\newblock URL \url{https://github.com/langchain-ai/langchain}.

\bibitem[Chen et~al.(2020)Chen, Kornblith, Swersky, Norouzi, and Hinton]{chen2020big}
T.~Chen, S.~Kornblith, K.~Swersky, M.~Norouzi, and G.~E. Hinton.
\newblock Big self-supervised models are strong semi-supervised learners.
\newblock \emph{Advances in neural information processing systems}, 2020.

\bibitem[Curth and van~der Schaar(2023)]{curth2023search}
A.~Curth and M.~van~der Schaar.
\newblock In search of insights, not magic bullets: Towards demystification of the model selection dilemma in heterogeneous treatment effect estimation.
\newblock \emph{arXiv preprint arXiv:2302.02923}, 2023.

\bibitem[Dallas et~al.(2017)Dallas, Machairas, and Papadopoulos]{dallas2017comparison}
S.~Dallas, K.~Machairas, and E.~Papadopoulos.
\newblock A comparison of ordinary differential equation solvers for dynamical systems with impacts.
\newblock \emph{Journal of Computational and Nonlinear Dynamics}, 2017.

\bibitem[Faller et~al.(2023)Faller, Vankadara, Mastakouri, Locatello, and Janzing]{faller2023self}
P.~M. Faller, L.~C. Vankadara, A.~A. Mastakouri, F.~Locatello, and D.~Janzing.
\newblock Self-compatibility: Evaluating causal discovery without ground truth.
\newblock \emph{arXiv preprint arXiv:2307.09552}, 2023.

\bibitem[Geffner et~al.(2022)Geffner, Antoran, Foster, Gong, Ma, Kiciman, Sharma, Lamb, Kukla, Pawlowski, Allamanis, and Zhang]{geffner2022deep}
T.~Geffner, J.~Antoran, A.~Foster, W.~Gong, C.~Ma, E.~Kiciman, A.~Sharma, A.~Lamb, M.~Kukla, N.~Pawlowski, M.~Allamanis, and C.~Zhang.
\newblock Deep end-to-end causal inference.
\newblock \emph{arXiv preprint arXiv:2202.02195}, 2022.

\bibitem[Gutierrez and Gérardy(2017)]{gutierrez17a}
P.~Gutierrez and J.-Y. Gérardy.
\newblock Causal inference and uplift modelling: A review of the literature.
\newblock In \emph{International Conference on Predictive Applications and APIs}. PMLR, 2017.

\bibitem[He et~al.(2021)He, Zhao, and Chu]{he2021automl}
X.~He, K.~Zhao, and X.~Chu.
\newblock Automl: A survey of the state-of-the-art.
\newblock \emph{Knowledge-Based Systems}, 2021.

\bibitem[Hutter et~al.(2019)Hutter, Kotthoff, and Vanschoren]{hutter2019automated}
F.~Hutter, L.~Kotthoff, and J.~Vanschoren.
\newblock \emph{Automated machine learning: methods, systems, challenges}.
\newblock Springer Nature, 2019.

\bibitem[Immer et~al.(2023)Immer, Schultheiss, Vogt, Sch{\"o}lkopf, B{\"u}hlmann, and Marx]{immer2023identifiability}
A.~Immer, C.~Schultheiss, J.~E. Vogt, B.~Sch{\"o}lkopf, P.~B{\"u}hlmann, and A.~Marx.
\newblock On the identifiability and estimation of causal location-scale noise models.
\newblock In \emph{International Conference on Machine Learning}, 2023.

\bibitem[Ke et~al.(2022)Ke, Chiappa, Wang, Goyal, Bornschein, Rey, Weber, Botvinic, Mozer, and Rezende]{ke2022learning}
N.~R. Ke, S.~Chiappa, J.~Wang, A.~Goyal, J.~Bornschein, M.~Rey, T.~Weber, M.~Botvinic, M.~Mozer, and D.~J. Rezende.
\newblock Learning to induce causal structure.
\newblock \emph{arXiv preprint arXiv:2204.04875}, 2022.

\bibitem[Ke et~al.(2023)Ke, Dunn, Bornschein, Chiappa, Rey, Lespiau, Cassirer, Wang, Weber, Barrett, et~al.]{ke2023discogen}
N.~R. Ke, S.-J. Dunn, J.~Bornschein, S.~Chiappa, M.~Rey, J.-B. Lespiau, A.~Cassirer, J.~Wang, T.~Weber, D.~Barrett, et~al.
\newblock Discogen: Learning to discover gene regulatory networks.
\newblock \emph{arXiv preprint arXiv:2304.05823}, 2023.

\bibitem[Keropyan et~al.(2023)Keropyan, Strieder, and Drton]{keropyan2023rank}
G.~Keropyan, D.~Strieder, and M.~Drton.
\newblock Rank-based causal discovery for post-nonlinear models.
\newblock In \emph{International Conference on Artificial Intelligence and Statistics}, 2023.

\bibitem[Kiciman et~al.(2022)Kiciman, Dillon, Edge, Foster, Hilmkil, Jennings, Ma, Ness, Pawlowski, Sharma, et~al.]{kiciman2022causal}
E.~Kiciman, E.~W. Dillon, D.~Edge, A.~Foster, A.~Hilmkil, J.~Jennings, C.~Ma, R.~Ness, N.~Pawlowski, A.~Sharma, et~al.
\newblock A causal ai suite for decision-making.
\newblock In \emph{NeurIPS 2022 Workshop on Causality for Real-world Impact}, 2022.

\bibitem[K{\i}c{\i}man et~al.(2023)K{\i}c{\i}man, Ness, Sharma, and Tan]{kiciman2023causal}
E.~K{\i}c{\i}man, R.~Ness, A.~Sharma, and C.~Tan.
\newblock Causal reasoning and large language models: Opening a new frontier for causality.
\newblock \emph{arXiv preprint arXiv:2305.00050}, 2023.

\bibitem[Kong et~al.(2023)Kong, Huang, Xie, Xing, Chi, and Zhang]{kong2023identification}
L.~Kong, B.~Huang, F.~Xie, E.~Xing, Y.~Chi, and K.~Zhang.
\newblock Identification of nonlinear latent hierarchical models.
\newblock \emph{arXiv preprint arXiv:2306.07916}, 2023.

\bibitem[Lachapelle et~al.(2019)Lachapelle, Brouillard, Deleu, and Lacoste-Julien]{lachapelle2019gradient}
S.~Lachapelle, P.~Brouillard, T.~Deleu, and S.~Lacoste-Julien.
\newblock Gradient-based neural dag learning.
\newblock \emph{arXiv preprint arXiv:1906.02226}, 2019.

\bibitem[Li et~al.(2020)Li, Xiao, and Tian]{li2020supervised}
H.~Li, Q.~Xiao, and J.~Tian.
\newblock Supervised whole dag causal discovery.
\newblock \emph{arXiv preprint arXiv:2006.04697}, 2020.

\bibitem[Liu et~al.(2010)Liu, Roeder, and Wasserman]{liu2010stability}
H.~Liu, K.~Roeder, and L.~Wasserman.
\newblock Stability approach to regularization selection (stars) for high dimensional graphical models.
\newblock \emph{Advances in neural information processing systems}, 2010.

\bibitem[Long et~al.(2023)Long, Pich{\'e}, Zantedeschi, Schuster, and Drouin]{long2023causal}
S.~Long, A.~Pich{\'e}, V.~Zantedeschi, T.~Schuster, and A.~Drouin.
\newblock Causal discovery with language models as imperfect experts.
\newblock \emph{arXiv preprint arXiv:2307.02390}, 2023.

\bibitem[Lorch et~al.(2022)Lorch, Sussex, Rothfuss, Krause, and Sch{\"o}lkopf]{lorch2022amortized}
L.~Lorch, S.~Sussex, J.~Rothfuss, A.~Krause, and B.~Sch{\"o}lkopf.
\newblock Amortized inference for causal structure learning.
\newblock \emph{Advances in Neural Information Processing Systems}, 2022.

\bibitem[Maathuis et~al.(2009)Maathuis, Kalisch, and B{\"u}hlmann]{maathuis2009est}
M.~H. Maathuis, M.~Kalisch, and P.~B{\"u}hlmann.
\newblock {Estimating high-dimensional intervention effects from observational data}.
\newblock \emph{The Annals of Statistics}, 2009.
\newblock \doi{10.1214/09-AOS685}.

\bibitem[Machlanski et~al.(2023)Machlanski, Samothrakis, and Clarke]{machlanski2023hyperparameter}
D.~Machlanski, S.~Samothrakis, and P.~Clarke.
\newblock Hyperparameter tuning and model evaluation in causal effect estimation.
\newblock \emph{arXiv preprint arXiv:2303.01412}, 2023.

\bibitem[Mahajan et~al.(2022)Mahajan, Mitliagkas, Neal, and Syrgkanis]{mahajan2022empirical}
D.~Mahajan, I.~Mitliagkas, B.~Neal, and V.~Syrgkanis.
\newblock Empirical analysis of model selection for heterogenous causal effect estimation.
\newblock \emph{arXiv preprint arXiv:2211.01939}, 2022.

\bibitem[Matthieu and Ga{\"e}l(2023)]{matthieu2023select}
D.~Matthieu and V.~Ga{\"e}l.
\newblock How to select predictive models for causal inference?
\newblock \emph{arXiv preprint arXiv:2302.00370}, 2023.

\bibitem[Mohd~Razali and Yap(2011)]{razali2011comparisions}
N.~Mohd~Razali and B.~Yap.
\newblock Power comparisons of shapiro-wilk, kolmogorov-smirnov, lilliefors and anderson-darling tests.
\newblock \emph{J. Stat. Model. Analytics}, 2, 01 2011.

\bibitem[Ouyang et~al.(2022)Ouyang, Wu, Jiang, Almeida, Wainwright, Mishkin, Zhang, Agarwal, Slama, Ray, et~al.]{ouyang2022training}
L.~Ouyang, J.~Wu, X.~Jiang, D.~Almeida, C.~Wainwright, P.~Mishkin, C.~Zhang, S.~Agarwal, K.~Slama, A.~Ray, et~al.
\newblock Training language models to follow instructions with human feedback.
\newblock \emph{Advances in Neural Information Processing Systems}, 35:\penalty0 27730--27744, 2022.

\bibitem[Peters et~al.(2014)Peters, Mooij, Janzing, and Sch{{\"o}}lkopf]{peters14a}
J.~Peters, J.~M. Mooij, D.~Janzing, and B.~Sch{{\"o}}lkopf.
\newblock Causal discovery with continuous additive noise models.
\newblock \emph{Journal of Machine Learning Research}, 2014.

\bibitem[Peters et~al.(2017)Peters, Janzing, and Sch{\"o}lkopf]{peters2017elements}
J.~Peters, D.~Janzing, and B.~Sch{\"o}lkopf.
\newblock \emph{Elements of causal inference: foundations and learning algorithms}.
\newblock The MIT Press, 2017.

\bibitem[Petersen et~al.(2022)Petersen, Ramsey, Ekstr{\o}m, and Spirtes]{petersen2022causal}
A.~H. Petersen, J.~Ramsey, C.~T. Ekstr{\o}m, and P.~Spirtes.
\newblock Causal discovery for observational sciences using supervised machine learning.
\newblock \emph{arXiv preprint arXiv:2202.12813}, 2022.

\bibitem[Pobrotyn et~al.(2020)Pobrotyn, Bartczak, Synowiec, Bialobrzeski, and Bojar]{Pobrotyn2020ContextAwareLT}
P.~Pobrotyn, T.~Bartczak, M.~Synowiec, R.~Bialobrzeski, and J.~Bojar.
\newblock Context-aware learning to rank with self-attention.
\newblock \emph{arXiv}, abs/2005.10084, 2020.

\bibitem[Postawa et~al.(2020)Postawa, Szczygie{\l}, and Ku{\l}a{\.z}y{\'n}ski]{postawa2020comprehensive}
K.~Postawa, J.~Szczygie{\l}, and M.~Ku{\l}a{\.z}y{\'n}ski.
\newblock A comprehensive comparison of ode solvers for biochemical problems.
\newblock \emph{Renewable Energy}, 2020.

\bibitem[Raghu et~al.(2018)Raghu, Poon, and Benos]{raghu2018evaluation}
V.~K. Raghu, A.~Poon, and P.~V. Benos.
\newblock Evaluation of causal structure learning methods on mixed data types.
\newblock In \emph{ACM SIGKDD Workshop on Causal Discovery}, 2018.

\bibitem[Raschka(2018)]{raschka2018model}
S.~Raschka.
\newblock Model evaluation, model selection, and algorithm selection in machine learning.
\newblock \emph{arXiv preprint arXiv:1811.12808}, 2018.

\bibitem[Rolling and Yang(2014)]{rolling2014model}
C.~A. Rolling and Y.~Yang.
\newblock Model selection for estimating treatment effects.
\newblock \emph{Journal of the Royal Statistical Society Series B: Statistical Methodology}, 2014.

\bibitem[Sachs et~al.(2005)Sachs, Perez, Pe'er, Lauffenburger, and Nolan]{sachs2005causal}
K.~Sachs, O.~Perez, D.~Pe'er, D.~A. Lauffenburger, and G.~P. Nolan.
\newblock Causal protein-signaling networks derived from multiparameter single-cell data.
\newblock \emph{Science}, 2005.

\bibitem[Saito and Yasui(2020)]{saito2020counterfactual}
Y.~Saito and S.~Yasui.
\newblock Counterfactual cross-validation: Stable model selection procedure for causal inference models.
\newblock In \emph{International Conference on Machine Learning}, 2020.

\bibitem[Schick et~al.(2023)Schick, Dwivedi-Yu, Dess{\`\i}, Raileanu, Lomeli, Zettlemoyer, Cancedda, and Scialom]{schick2023toolformer}
T.~Schick, J.~Dwivedi-Yu, R.~Dess{\`\i}, R.~Raileanu, M.~Lomeli, L.~Zettlemoyer, N.~Cancedda, and T.~Scialom.
\newblock Toolformer: Language models can teach themselves to use tools.
\newblock \emph{arXiv preprint arXiv:2302.04761}, 2023.

\bibitem[Schuler et~al.(2018)Schuler, Baiocchi, Tibshirani, and Shah]{schuler2018comparison}
A.~Schuler, M.~Baiocchi, R.~Tibshirani, and N.~Shah.
\newblock A comparison of methods for model selection when estimating individual treatment effects.
\newblock \emph{arXiv preprint arXiv:1804.05146}, 2018.

\bibitem[Scutari(2010)]{scutari2010learn}
M.~Scutari.
\newblock Learning bayesian networks with the {bnlearn} {R} package.
\newblock \emph{Journal of Statistical Software}, 2010.
\newblock \doi{10.18637/jss.v035.i03}.

\bibitem[Scutari et~al.(2014)Scutari, Howell, Balding, and Mackay]{scutari2014multiple}
M.~Scutari, P.~Howell, D.~J. Balding, and I.~Mackay.
\newblock Multiple quantitative trait analysis using bayesian networks.
\newblock \emph{Genetics}, 2014.

\bibitem[Shimizu et~al.(2011)Shimizu, Inazumi, Sogawa, Hyvarinen, Kawahara, Washio, Hoyer, Bollen, and Hoyer]{shimizu2011directlingam}
S.~Shimizu, T.~Inazumi, Y.~Sogawa, A.~Hyvarinen, Y.~Kawahara, T.~Washio, P.~O. Hoyer, K.~Bollen, and P.~Hoyer.
\newblock Directlingam: A direct method for learning a linear non-gaussian structural equation model.
\newblock \emph{Journal of Machine Learning Research (JMLR)}, 2011.

\bibitem[Squires and Uhler(2022)]{squires2022causal}
C.~Squires and C.~Uhler.
\newblock Causal structure learning: a combinatorial perspective.
\newblock \emph{arXiv preprint arXiv:2206.01152}, 2022.

\bibitem[Strobl(2021)]{strobl2021automated}
E.~V. Strobl.
\newblock Automated hyperparameter selection for the pc algorithm.
\newblock \emph{Pattern Recognition Letters}, 2021.

\bibitem[Uemura et~al.(2022)Uemura, Takagi, Takayuki, Yoshida, and Shimizu]{uemura2022multivariate}
K.~Uemura, T.~Takagi, K.~Takayuki, H.~Yoshida, and S.~Shimizu.
\newblock A multivariate causal discovery based on post-nonlinear model.
\newblock In \emph{Conference on Causal Learning and Reasoning}, 2022.

\bibitem[Van~den Bulcke et~al.(2006)Van~den Bulcke, Van~Leemput, Naudts, van Remortel, Ma, Verschoren, De~Moor, and Marchal]{van2006syntren}
T.~Van~den Bulcke, K.~Van~Leemput, B.~Naudts, P.~van Remortel, H.~Ma, A.~Verschoren, B.~De~Moor, and K.~Marchal.
\newblock Syntren: a generator of synthetic gene expression data for design and analysis of structure learning algorithms.
\newblock \emph{BMC bioinformatics}, 2006.

\bibitem[Wang and Kording(2022)]{wang2022meta}
X.~Wang and K.~Kording.
\newblock Meta-learning causal discovery.
\newblock \emph{arXiv preprint arXiv:2209.05598}, 2022.

\bibitem[Wang et~al.(2018)Wang, Li, Golbandi, Bendersky, and Najork]{wang2018lambdaloss}
X.~Wang, C.~Li, N.~Golbandi, M.~Bendersky, and M.~Najork.
\newblock The lambdaloss framework for ranking metric optimization.
\newblock In \emph{Proceedings of the 27th ACM international conference on information and knowledge management}, pages 1313--1322, 2018.

\bibitem[Wei et~al.(2022)Wei, Tay, Bommasani, Raffel, Zoph, Borgeaud, Yogatama, Bosma, Zhou, Metzler, Chi, Hashimoto, Vinyals, Liang, Dean, and Fedus]{wei2022emergent}
J.~Wei, Y.~Tay, R.~Bommasani, C.~Raffel, B.~Zoph, S.~Borgeaud, D.~Yogatama, M.~Bosma, D.~Zhou, D.~Metzler, E.~H. Chi, T.~Hashimoto, O.~Vinyals, P.~Liang, J.~Dean, and W.~Fedus.
\newblock Emergent abilities of large language models.
\newblock \emph{Transactions on Machine Learning Research}, 2022.

\bibitem[Yao et~al.(2021)Yao, Chu, Li, Li, Gao, and Zhang]{yao2021survey}
L.~Yao, Z.~Chu, S.~Li, Y.~Li, J.~Gao, and A.~Zhang.
\newblock A survey on causal inference.
\newblock \emph{ACM Transactions on Knowledge Discovery from Data (TKDD)}, 2021.

\bibitem[Yu et~al.(2019)Yu, Chen, Gao, and Yu]{yu2019dag}
Y.~Yu, J.~Chen, T.~Gao, and M.~Yu.
\newblock Dag-gnn: Dag structure learning with graph neural networks.
\newblock \emph{International Conference on Machine Learning}, 2019.

\bibitem[Ze{v{c}}evi{\'c} et~al.(2023)Ze{v{c}}evi{\'c}, Willig, Dhami, and Kersting]{zevcevic2023causal}
M.~Ze{v{c}}evi{\'c}, M.~Willig, D.~S. Dhami, and K.~Kersting.
\newblock Causal parrots: Large language models may talk causality but are not causal.
\newblock \emph{Transactions on Machine Learning Research}, 2023.

\bibitem[Zhai et~al.(2019)Zhai, Oliver, Kolesnikov, and Beyer]{zhai2019s4l}
X.~Zhai, A.~Oliver, A.~Kolesnikov, and L.~Beyer.
\newblock S4l: Self-supervised semi-supervised learning.
\newblock In \emph{Proceedings of the IEEE/CVF international conference on computer vision}, 2019.

\bibitem[Zhang et~al.(2023)Zhang, Bauer, Bennett, Gao, Gong, Hilmkil, Jennings, Ma, Minka, Pawlowski, et~al.]{zhang2023understanding}
C.~Zhang, S.~Bauer, P.~Bennett, J.~Gao, W.~Gong, A.~Hilmkil, J.~Jennings, C.~Ma, T.~Minka, N.~Pawlowski, et~al.
\newblock Understanding causality with large language models: Feasibility and opportunities.
\newblock \emph{arXiv preprint arXiv:2304.05524}, 2023.

\bibitem[Zhang and Hyvarinen(2012)]{zhang2012identifiability}
K.~Zhang and A.~Hyvarinen.
\newblock On the identifiability of the post-nonlinear causal model.
\newblock \emph{arXiv preprint arXiv:1205.2599}, 2012.

\bibitem[Zhang et~al.(2021)Zhang, Zhu, Kalander, Ng, Ye, Chen, and Pan]{zhang2021gcastle}
K.~Zhang, S.~Zhu, M.~Kalander, I.~Ng, J.~Ye, Z.~Chen, and L.~Pan.
\newblock gcastle: A python toolbox for causal discovery.
\newblock \emph{arXiv preprint arXiv:2111.15155}, 2021.

\bibitem[Zheng et~al.(2018)Zheng, Aragam, Ravikumar, and Xing]{zheng2018dags}
X.~Zheng, B.~Aragam, P.~K. Ravikumar, and E.~P. Xing.
\newblock Dags with no tears: Continuous optimization for structure learning.
\newblock \emph{Advances in neural information processing systems}, 2018.

\bibitem[Zheng et~al.(2020)Zheng, Dan, Aragam, Ravikumar, and Xing]{zheng2020learning}
X.~Zheng, C.~Dan, B.~Aragam, P.~Ravikumar, and E.~Xing.
\newblock Learning sparse nonparametric dags.
\newblock \emph{International Conference on Artificial Intelligence and Statistics}, 2020.

\end{thebibliography}

\clearpage

\appendix
\section{Additional details on synthetic data generation} \label{apdx:synthetic-data-details}

\begin{figure}
\centering
\includegraphics[scale=0.50]{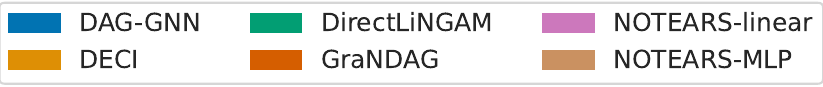}
\\
\vspace{5pt}
\subfigure[All Datasets]{
\includegraphics[scale=0.47]{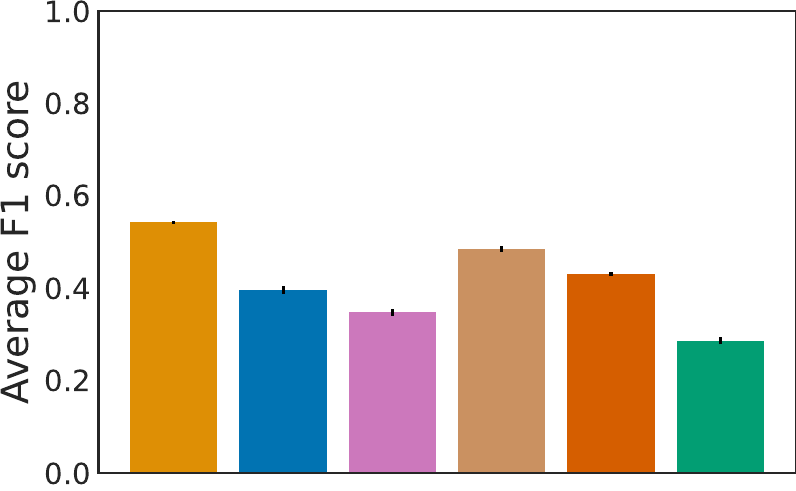}
}
\hspace{20pt}
\subfigure[Linear Gaussian datasets]{
\includegraphics[scale=0.47]{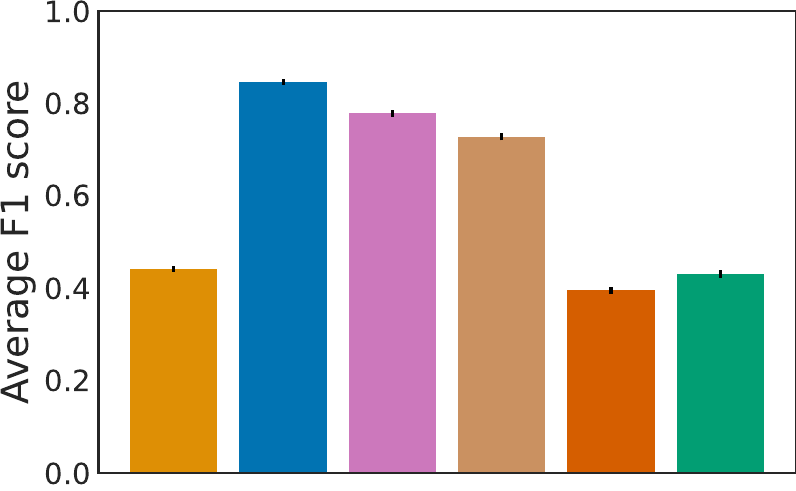}
}
\\
\vspace{5pt}
\subfigure[Linear non-Gaussian datasets]{
\includegraphics[scale=0.47]{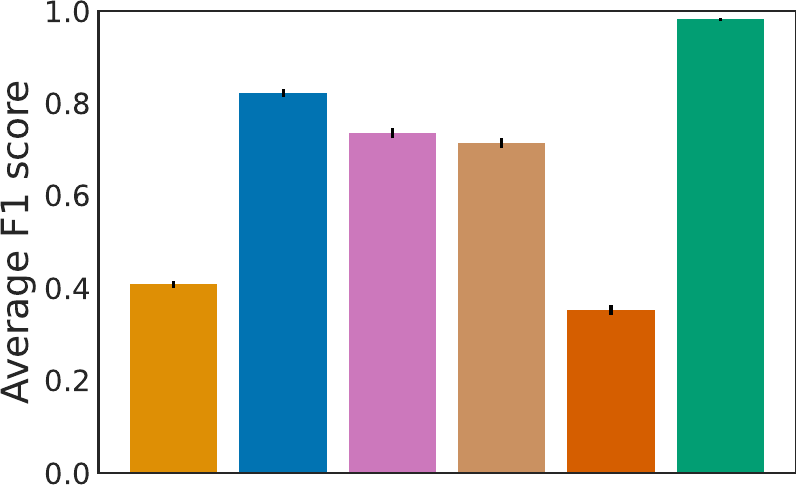}
}
\hspace{20pt}
\subfigure[Nonlinear datasets]{
\includegraphics[scale=0.47]{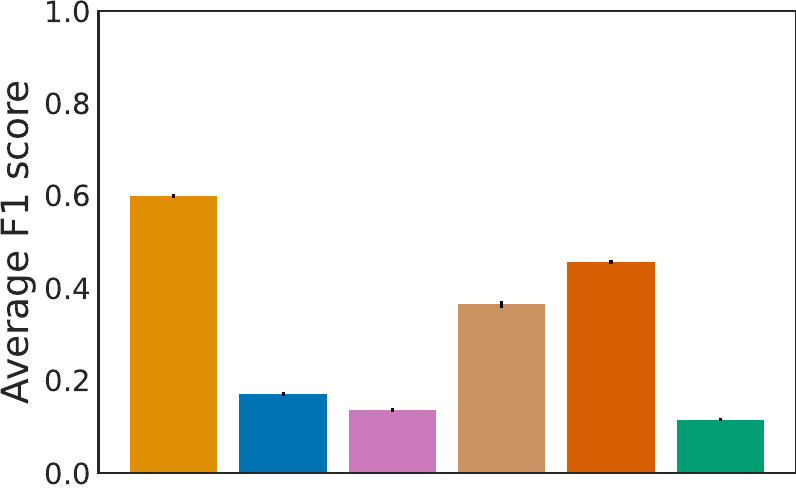}
}
\caption{The average F1 scores (error bars denote 95\% CIs) 
of the six causal discovery methods on synthetic data across 
(a) All datasets, and (b)-(d) different subsets of the data sliced according to
the underlying assumptions.}
\label{fig:apdx-synthetic-data-scores}
\end{figure}

In this section, we provide  additional details 
on the data-generating processes (DGP)
for the synthetic datasets (see Sec.~\ref{sec:results-synthetic-data-gen}).
Throughout, the graphs are sampled from an Erdos-Renyi 
distribution with the edge probabilities uniformly sampled
from $[0.3, 0.7]$.
Recall that each node $X_j$ is generated via the structural equation
$X_j := f_j( X_{\text{pa}(j; G_*)}, n_j )$, 
where $n_j$ is an exogenous noise term.
Unless stated otherwise, we use Gaussian noise with random variance: 
$n_j \sim \NM(0, \sigma^2_j)$ where $\sigma_j \sim \text{Uniform}([0.2, 2])$.
We randomly generate the following SCMs:
\begin{enumerate}
\item \emph{Linear Gaussian}: We simulate $X_j = w^\top_j X_{\text{pa}(j)} + n_j$,
where the coefficients $w_j \sim \\ \text{Uniform}( [-2, -0.5] \cup [0.5, 2] )$
(similar to \citet{zheng2018dags}).

\item \emph{Linear non-Gaussian}: This is the same as \emph{Linear Gaussian} but the noise $n_j$ has a uniform or exponential
distribution. Amongst the linear non-Gaussian SCMs, we use a $50$--$50\%$
split between uniformly and exponentially distributed noise.

\item \emph{Nonlinear Additive Noise Models (ANM)} \citep{peters14a}: We simulate $X_j = f_j(X_{\text{pa}(j)}) + n_j$, 
where $n_j$ is Gaussian and each $f_j$ is one of two nonlinear functions 
(similar to \citet{zheng2020learning}):
(i) random function from a Gaussian Process (GP), or
(ii) (Additive GP) $f_i(X_{\text{Pa}(i)}) = \sum_{j \in \text{Pa}(i))} g_j(X_j)$, where
each $g_j$ is a random function from a GP.
In both cases, we use a GP with an RBF kernel with scale $1$.

\item \emph{Post-nonlinear (PNL) model} \citep{zhang2012identifiability}:
We simulate $X_j = f_j(g_j(X_{\text{pa}(j)} + n_j))$ where $g_i, f_i$
are nonlinear functions from one of the following PNL models:
(i) $f_i$ and $g_i$ are sampled as weighted sums of GPs and sigmoids )(we use the same DGP as \citet{uemura2022multivariate}), or
(ii) $f_i$ is a polynomial and $g_i$ is the cube-root (we use the same DGP as \citet{keropyan2023rank});

\item \emph{Location-scale model} \citep{immer2023identifiability}: This is a heteroskedastic noise model with
$X_j = f_j(X_{\text{pa}(j)}) + g_j(X_{\text{pa}(j)}) \cdot n_j$,
where $n_j$ is Gaussian and $f_j, g_j$ are random functions from a GP with an RBF kernel with scale $1$.
\end{enumerate}

We also compare the scores of the six causal discovery methods (Table~\ref{table:candidate-causal-discovery-methods})
across the all datasets and various subsets thereof (see Fig.~\ref{fig:apdx-synthetic-data-scores}).
We see that \emph{DECI} performs the best overall, but across different subsets,
different methods have the highest average F1 scores (e.g., DirectLiNGAM is the best on
average on Linear non-Gaussian datasets).

\section{Rank-based loss functions} \label{apdx:rank-based-losses}

\begin{figure}
\centering
\includegraphics[scale=0.43]{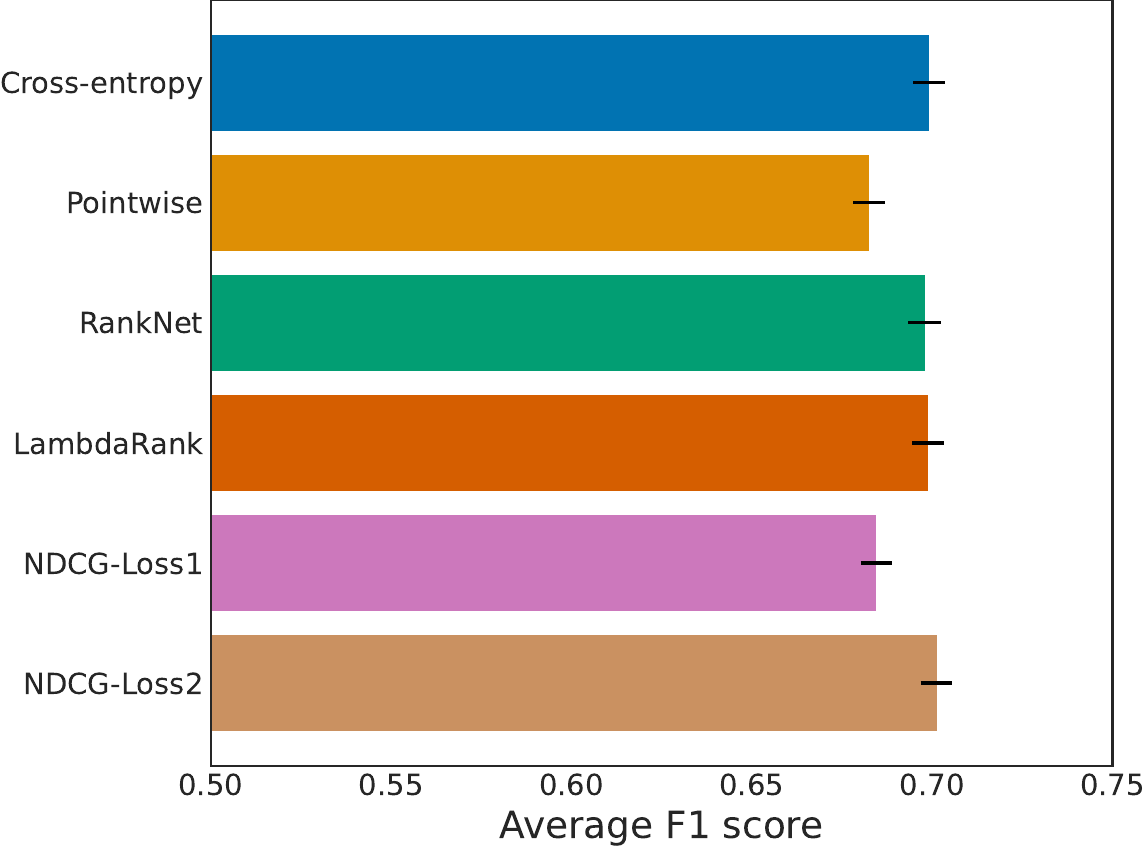}
\caption{Average F1 scores of the various rank-based loss functions on the test set. 
We do not observe a significant improvement relative to the cross-entropy loss.}
\label{fig:apdx-rank-losses}
\end{figure}

In the training data, we also have access to the raw scores for all candidate method 
on every dataset. Thus, instead of using a cross-entropy loss, it is also possible to
treat the prediction problem as a learning-to-rank problem and apply 
rank-based loss functions (we refer the reader to \citet{wang2018lambdaloss} for an overview
of rank-based loss functions). We compare the cross-entropy to various rank-based loss functions
(see Fig.~\ref{fig:apdx-rank-losses}) but find no significant improvement for our task.
We use the \emph{allrank} \citep{Pobrotyn2020ContextAwareLT} \footnote{We use the code from {\small \url{https://github.com/allegro/allRank}}.} and test the following rank-based losses:
(i) \emph{Pointwise} (where we directly regress the F1-scores for each candidate method),
(ii) \emph{RankNet} \citep{burges2005learning}, 
(iii) \emph{LambdaRank} \citep{burges2010ranknet},
and (iv) \emph{NDCG-Loss1} and \emph{NDCG-Loss2} as described in \citet[Sec.~5.2]{wang2018lambdaloss}.

\section{Additional results on the semi-synthetic and real-world benchmarks} \label{apdx:real-world}

\begin{figure}
\centering
\includegraphics[scale=0.50]{figures/real_world_appendix_legend.pdf}
\\
\vspace{5pt}
\subfigure[MAGIC-NIAB]{
\includegraphics[scale=0.45]{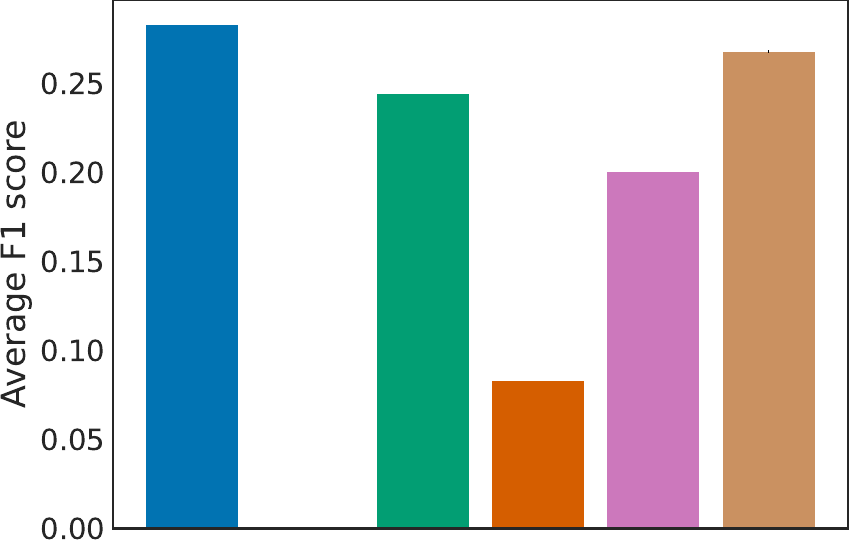}
\label{fig:apdx-results-real-world-magic-niab}
}
\hfill
\subfigure[MAGIC-IRRI]{
\includegraphics[scale=0.45]{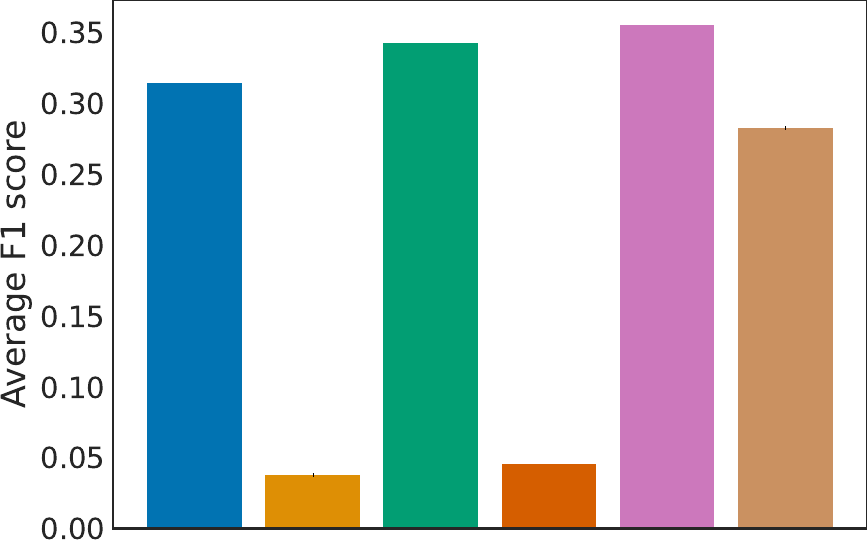}
\label{fig:apdx-results-real-world-magic-irri}
}
\\
\vspace{5pt}
\subfigure[SynTReN]{
\includegraphics[scale=0.45]{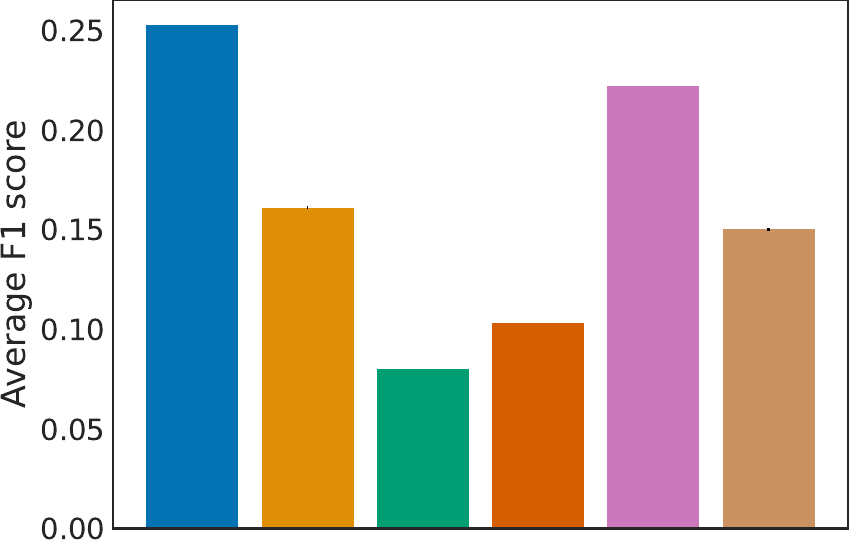}
\label{fig:apdx-results-real-world-syntren}
}
\hfill
\subfigure[Protein cells]{
\includegraphics[scale=0.45]{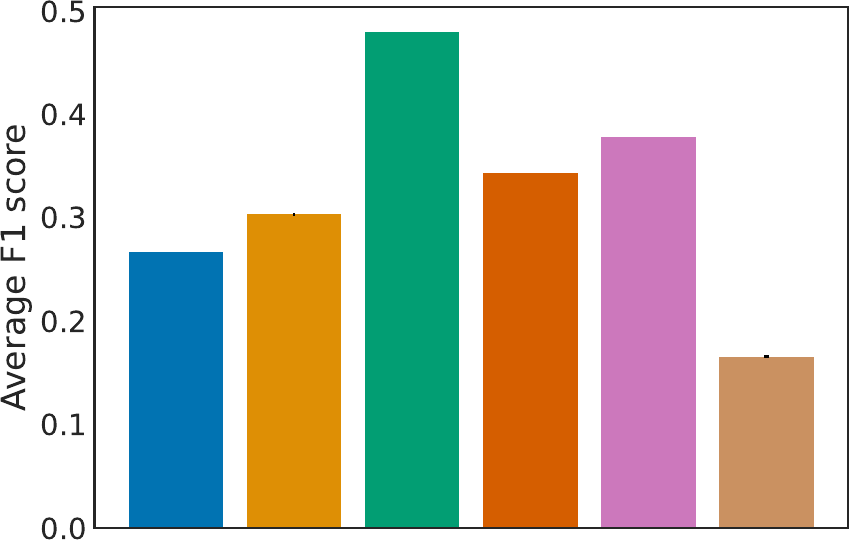}
\label{fig:apdx-results-real-world-protein}
}
\caption{The F1 scores of the six causal discovery methods (averaged over $10$ runs of each method) 
on the four semi-synthetic and real-world benchmarks (for \emph{MAGIC-NIAB}, the bar for DECI is missing
because it outputs a nearly empty graph resulting in a zero F1 score).}
\label{fig:apdx-results-real-world}
\end{figure}

For the four semi-synthetic and real-world benchmarks (see Sec.~\ref{sec:results-real-world}),
we compute the F1-score for each candidate method by averaging the
scores over $10$ runs (see Fig.~\ref{fig:apdx-results-real-world} for the scores of each method on the four benchmarks).
Next, we also show the predicted rankings of the six methods from
\emph{CAMP-SemiSup} and \emph{CAMP-Sup} for each of the four benchmarks
(see Table~\ref{tab:apdx-real-world-rankings}).

\begin{table}
\centering
\begin{tabular}{ll}
\toprule
\textbf{Benchmarks} & \textbf{Predicted ranking (from best to worst)}  \\
\midrule
 \multicolumn{2}{l}{\textbf{MAGIC-NIAB}} \vspace{5pt} \\
 Oracle & DAG-GNN, NOTEARS-MLP, DirectLiNGAM, NOTEARS-linear, GranDAG, DECI  \\
 CAMP-SemiSup & DAG-GNN, NOTEARS-linear, NOTEARS-MLP, DECI, DirectLiNGAM, GranDAG  \\
 CAMP-Sup & DECI, NOTEARS-MLP, GranDAG, DAG-GNN, NOTEARS-linear, DirectLiNGAM  \\
 \midrule
 \multicolumn{2}{l}{\textbf{MAGIC-IRRI}} \vspace{5pt} \\
 Oracle & NOTEARS-linear, DirectLiNGAM, DAG-GNN, NOTEARS-MLP, GranDAG, DECI  \\
 CAMP-SemiSup & DAG-GNN, NOTEARS-linear, NOTEARS-MLP, DECI, DirectLiNGAM, GranDAG  \\
 CAMP-Sup & DECI, NOTEARS-MLP, DAG-GNN, GranDAG, NOTEARS-linear, DirectLiNGAM  \\
 \midrule
 \multicolumn{2}{l}{\textbf{SynTReN}} \vspace{5pt} \\
 Oracle & DAG-GNN, NOTEARS-linear, DECI, NOTEARS-MLP, GranDAG, DirectLiNGAM  \\
 CAMP-SemiSup & DAG-GNN, NOTEARS-linear, NOTEARS-MLP, DirectLiNGAM, GranDAG, DECI  \\
 CAMP-Sup & DAG-GNN, NOTEARS-linear, NOTEARS-MLP, DirectLiNGAM, DECI, GranDAG  \\
\midrule
 \multicolumn{2}{l}{\textbf{Protein cells}} \vspace{5pt} \\
 Oracle & DirectLiNGAM, NOTEARS-linear, GranDAG, DECI, DAG-GNN, NOTEARS-MLP  \\
 CAMP-SemiSup & DirectLiNGAM, NOTEARS-linear, DAG-GNN, NOTEARS-MLP, GranDAG, DECI  \\
 CAMP-Sup & DirectLiNGAM, DAG-GNN, NOTEARS-linear, NOTEARS-MLP, DECI, GranDAG  \\
 \bottomrule
\end{tabular}
\caption{The predicted rankings of the six causal discovery methods from \emph{CAMP-SemiSup} and \emph{CAMP-Sup}
for the four semi-synthetic and real-world benchmarks (also see Fig.~\ref{fig:apdx-results-real-world} for the F1-scores of the various methods on each benchmark).}
\label{tab:apdx-real-world-rankings}
\end{table}

\end{document}